\documentclass[10pt,twocolumn,letterpaper]{article}

\usepackage{cvpr}              

\usepackage[table]{xcolor}

\definecolor{cvprblue}{rgb}{0.21,0.49,0.74}
\usepackage[pagebackref,breaklinks,colorlinks,citecolor=cvprblue]{hyperref}
\usepackage{booktabs}
\usepackage{multirow}
\usepackage{multicol}
\usepackage{graphicx}

\usepackage{pifont}
\newcommand{\cmark}{\ding{51}}%

\def\paperID{13011} 
\def\confName{CVPR}
\def\confYear{2024}

\newcommand{\model}{\textsc{Groundhog}}
\newcommand{\icon}{\inlinegraphics{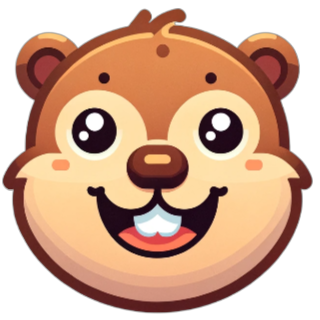}}

\usepackage{graphicx,calc}
\newlength\myheight
\newlength\mydepth
\settototalheight\myheight{Xygp}
\newcommand*\inlinegraphics[1]{%
  \settototalheight\myheight{Xygp}%
  \settodepth\mydepth{Xygp}%
  \raisebox{-.25\mydepth}{\includegraphics[height=1.2\myheight]{#1}}%
}

\title{\vspace*{-10pt}\model~\icon: Grounding Large Language Models to Holistic Segmentation\vspace*{-10pt}}

\author{
    Yichi Zhang\textsuperscript{\rm 1}\stepcounter{footnote}\thanks{~Work done during internship at Amazon AGI.}~,
    Ziqiao Ma\textsuperscript{\rm 1}\footnotemark[2]~, 
    Xiaofeng Gao\textsuperscript{\rm 2},
    Suhaila Shakiah\textsuperscript{\rm 2},
    Qiaozi Gao\textsuperscript{\rm 2},
    Joyce Chai\textsuperscript{\rm 1}
    \\
    \textsuperscript{\rm 1}University of Michigan, \textsuperscript{\rm 2}Amazon AGI \\
    {\tt zhangyic@umich.edu} \\
    {\tt \url{https://groundhog-mllm.github.io/}}
}

\begin{document}

\input{vis/fig_teaser}
\maketitle
\begin{abstract}

\vspace*{-10pt}

Most multimodal large language models (MLLMs) learn language-to-object grounding through causal language modeling where grounded objects are captured by bounding boxes as sequences of location tokens.
This paradigm lacks pixel-level representations that are important for fine-grained visual understanding and diagnosis. 
In this work, we introduce \model, an MLLM developed by \underline{ground}ing Large Language Models to \underline{ho}listic se\underline{g}mentation. 
\model~incorporates a masked feature extractor and converts extracted features into visual entity tokens for the MLLM backbone, which then connects groundable phrases to unified grounding masks by retrieving and merging the entity masks.
To train \model, we carefully curated M3G2, a grounded visual instruction tuning dataset with \underline{M}ulti-\underline{M}odal \underline{M}ulti-\underline{G}rained \underline{G}rounding, by harvesting a collection of segmentation-grounded datasets with rich annotations.
Our experimental results show that \model~achieves superior performance on various language grounding tasks without task-specific fine-tuning, and significantly reduces object hallucination.
\model~also demonstrates better grounding towards complex forms of visual input and provides easy-to-understand diagnosis in failure cases. 

\end{abstract}
\vspace{-15pt}
\section{Introduction}
\label{sec:intro}

Multimodal large language models (MLLMs) have received an increasing amount of attention to address tasks that necessitate non-linguistic knowledge, e.g., perception and reasoning about the visual world~\cite{yin2023survey,li2023multimodal}. 
For fine-grained visual understanding, grounded MLLMs often learn language-to-object grounding by causal language modeling, where grounded objects are captured by bounding boxes as sequences of location tokens.
However, bounding boxes are insufficient in indicating amorphous stuff~\cite{caesar2018coco}, semantic parts of objects~\cite{gonzalez2018objects}, finer-grained regions with irregular shapes~\cite{han2022vision}, or groups of instances at the same time.
As a result, a single bounding box can often include other irrelevant semantics in order to engulf the target entities, leading to ambiguity in detection. 
In addition, the generated box coordinate lacks interpretability. 
When the model hallucinates, such as incorrectly predicting the association between objects and language, it is hard to diagnose whether the problem is due to the model's failure to detect the object, or its incorrect alignment of the object with language.

\begin{figure*}[t!]
    \centering
    \includegraphics[width=1.0\linewidth]{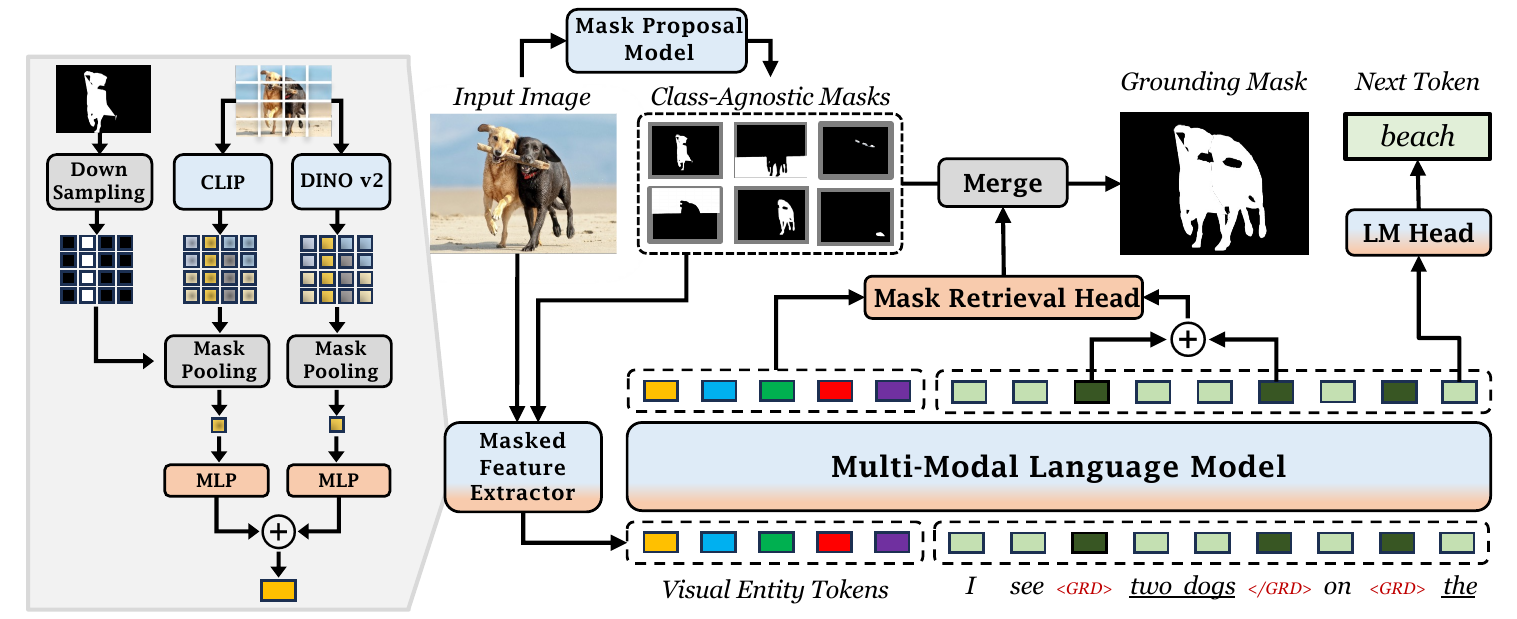}
    \vspace{-20pt}
    \caption{The model architecture of \model~model. Given a set of class-agnostic entity mask proposals, the masked feature extractor first extracts the feature of each entity as the visual input of the multi-modal large language model (left). The output hidden states of the grounding tokens are averaged and used to retrieve the entities to ground, which will be merged into a single grounding mask for the phrase. Modules are colored by their trainability: 
    parameter-free operators (grey), frozen (blue), trainable (orange), and partially trainable (mix).\vspace{-12pt}}
    \label{fig:model}
\end{figure*}

To address these issues, in this work, we introduce \model, an MLLM developed by \underline{ground}ing Large Language Models to \underline{ho}listic se\underline{g}mentation. 
Our goal of language grounding is to connect text spans that refer to or can be deduced from visual information, termed as \textit{groundable phrases}~\cite{ma2023world}, to their corresponding regions of visual entities. 
\model~incorporates a masked feature extractor that takes an input image and a set of class-agnostic entity mask proposals, and converts each mask's features into visual entity tokens for an MLLM backbone. 
This MLLM then connects groundable phrases to unified grounding masks by retrieving and merging the entity masks.
Compared to previous grounded MLLMs, \model~unlocks unprecedented pixel-level vision-language alignment.
It naturally supports visual pointers as input, and can plug-in-and-play with any choice of mask proposal networks, e.g., Segment Anything Model (SAM)~\cite{kirillov2023segment}, domain-specific semantic segmentation models, or user-provided mask candidates.
We introduce an enhanced Mask2Former~\cite{cheng2022masked} as our default mask proposal network, which detects regions at multiple granularities, e.g., instances (things and stuff), semantic parts, and visual text, leading to a holistic coverage of visual semantics.

To train \model, we curated a \underline{M}ulti-\underline{M}odal \underline{M}ulti-\underline{G}rained \underline{G}rounding (M3G2) dataset consisting of 2.5M text-image pairs for visually grounded instruction tuning, consisting of 36 sub-problems derived and augmented from 27 existing datasets.
We present extensive experiments on vision-language tasks that require grounding, including grounded language generation with minimal object hallucination, language-guided segmentation, visual question answering with answer grounding, and referential dialog with spatial pointer inputs (Figure \ref{fig:demo}). 
Our empirical results show that \model, without task-specific fine-tuning, can achieve superior or comparable performance with previous models that either require fine-tuning or are specialized only for that dataset. 
In addition, \model~has supports easy-to-understand diagnosis when grounding fails.

\section{Our Method: \model}
\label{sec:model}

The language grounding task can be succinctly delineated into two fundamental components: \textit{localization} and \textit{recognition}, as established in the literature \citep{singh2018r, zhong2022regionclip, ma2023world}. 
Such categorization not only aids in the identification of object presence (objectness) without reliance on specific object classes, but also sets the stage for models to be robust in open-vocabulary settings. 
Building upon this framework, we formulate the grounding process as an \textit{entity segment selection} problem, 
which involves (1) proposing entity segmentation masks where the masks encapsulate regions with discernible semantic content, and (2) recognizing the retrieved entities through the understanding of both visual and language context. Concurrently performing both tasks is where MLLMs bring a distinct advantage.
This decoupled design of entity mask proposal and language-guided grounding brings several advantages. 
First, it allows independent improvement of the mask proposal model and MLLM, where specialized data, training, and inference setups can be applied. 
Second, by decoupling language grounding, it becomes straightforward to determine if a failure is due to the model's inability to propose the entity segment, or its misalignment of the object with the language, thus improving the interpretability of the whole framework. 
Third, as shown later, when connecting the two parts to work in tandem in a model-independent manner, the MLLM can benefit from multiple different vision specialist models in a plug-and-play fashion. 
In the remainder of this section, we give details of our model design.

\subsection{Building Entity Features from Masks}
\label{subsec:extractor}

Our approach assumes the availability of a mask proposal model, which is capable of generating a set of class-agnostic entity masks from an image with high coverage. 
In contrast to prior studies that relied on low-level features~\cite{liu2023visual,peng2023kosmos,chen2023shikra,dai2023instructblip}, \model~interprets the image as a collection of entities. 
The primary challenge then becomes the derivation of effective visual features to accurately represent these entities. 
To achieve a complete decoupling of the MLLM from the mask proposal model responsible for providing the masks, we propose to condition the entity features solely on the binary masks without using any embeddings from the mask proposal model. 
Specifically, the mask corresponding to each entity is employed to extract patch features from pretrained vision foundation models, such as CLIP~\cite{radford2021clip} and DINOv2~\cite{oquab2023dinov2}, through a convolutional mask pooling layer~\cite{dai2015convolutional}. 
Given that the feature map dimensions are usually smaller than those of the mask proposals, we resize the masks to match the size of the feature maps prior to pooling. 
The pooled features are then fed into a Multi-Layer Perceptron (MLP) network to align with the input embeddings of the MLLM. 
We empirically find the combination of CLIP and DINOv2 features yields the best result, and these features are added to obtain the final input visual entity tokens to the MLLM. 

\begin{figure}[!t]
    \centering
    \includegraphics[width=1.0\linewidth]{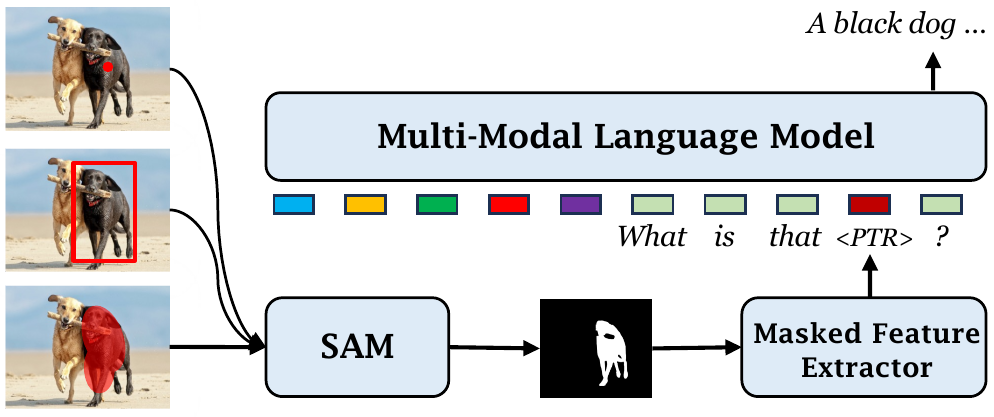}
    \vspace{-20pt}
    \caption{\model~can take arbitrary spatial prompts that can be resolved by an interactive segmentation model, such as SAM. The placeholder pointer token \texttt{<PTR>} will be replaced by the extracted entity features and fed as input to the model. \vspace{-15pt}}
    \label{fig:model-ptr}
\end{figure}

\vspace*{-10pt}
\paragraph{Spatial Prompts}
Furthermore, for grounded MLLMs to be more broadly applicable, they must be capable of interpreting multi-modal user inputs, including spatial prompts. 
Thanks to the mask model agnostic design, \model can seamlessly support such inputs. 
As demonstrated in Figure~\ref{fig:model-ptr}, by applying an interactive segmentation model such as Segment-Anything (SAM)~\cite{kirillov2023segment}, arbitrary spatial prompts can be translated into binary masks and processed by the same masked feature extractor we just introduced. 
This extracted feature for the pointed entity will replace the pointer token \texttt{<PTR>} placeholder in the textual input. 

\subsection{Language Grounding to Entity Segmentation}
\label{subsec:grounding}

Existing box-grounded MLLMs typically append location tokens after the groundable phrases~\cite{chen2021pix2seq,peng2023kosmos,chen2023shikra,you2023ferret}. 
However, this method is not readily interpretable. 
To alleviate this disconnect, we introduce a pair of grounding tokens \texttt{<GRD>} and \texttt{</GRD>} to indicate the start and end of groundable phrases, with the assumption that grounding these phrases requires mapping to certain representations of visual entities irrespective of the visual modality.
In Figure~\ref{fig:model}, a sentence can be represented as \texttt{I see <GRD> two dogs </GRD> on <GRD> the beach </GRD>}, with two distinct visual entities grounded. 
The representation of each groundable phrase, termed as the \textit{grounding query}, is obtained by adding \texttt{<GRD>} and \texttt{</GRD>}'s output embedding from the last transformer layer of the MLLM. 
The representation is then used to retrieve the entities that the phrase should be grounded to. 
In particular, we concatenate the grounding query with the last layer output of each visual entity token, and use an MLP to predict a scalar score for each entity. 
Finally, we merge all the mask proposals into one single mask with pixel-wise maximization:
\begin{equation*}
\vspace*{-3pt}
\mathcal{M}_{h,w} = \max_q \left( \mathcal{S}_q \cdot \widehat{\mathcal{M}}_{q,h,w} \right)\
\vspace*{-2pt}
\end{equation*}
where $\mathcal{S}_q$ is the normalized score of the $q$-th mask ranging from 0 to 1, and $\widehat{\mathcal{M}}_{q,h,w}$ denotes the pixel probability at position $(h, w)$ for the $q$-th mask.
Note that a phrase may ground to multiple entities, thus multiple mask proposals may get a high score simultaneously and be selected in conjunction.
One of the primary benefits of this decoupled design is its transparency in the selection of entities. 
Users can easily visualize both the mask proposals and their respective scores, providing a clear understanding of how a grounding mask is predicted. 
This level of clarity and interpretability is a significant advantage, offering users a tangible insight into the model's grounding process.

\subsection{Towards Holistic Entity Mask Proposals}
\label{subsec:proposal}

In order to support holistic language grounding to arbitrary segmentations, the entity proposal should have two essential properties. 
First, the proposals should strike a delicate balance in terms of semantic atomicity. 
While it is possible to merge multiple proposals later to form multi-entity segmentations, the reverse, i.e., dividing a single proposal into smaller segments, is not feasible. 
Therefore, instance segmentation is generally preferred over semantic segmentation. 
However, the segmentation should not be excessively fine-grained to the extent that it compromises basic semantic integrity. 
Over-segmentation can lead to a loss of the coherent concept of an entity, which is detrimental to the grounding process. 
Second, the entity proposals should have a high coverage of entities, encompassing a diverse range of granularities. This includes not only tangible objects (things) and amorphous concepts (stuff) but also extends to sub-components of objects (parts of things) and structured regions such as areas containing visual text.
The ability to propose entities across this spectrum of granularity is pivotal, as it directly determines the upper bound of the grounding capability of MLLM.

We initiated our study with a Mask2Former model pre-trained on the COCO panoptic segmentation dataset, capable of segmenting 134 object categories. 
However, preliminary experiments revealed its limitations in semantic coverage and adaptability to open-world scenarios. 
To enhance this, we developed Mask2Former+, an upgraded version designed for multi-grained segmentation. 
This upgrade involved creating a diverse dataset by merging annotations from various sources, including COCO~\cite{caesar2018coco}, LVIS~\cite{gupta2019lvis}, Entity-v2~\cite{qi2023high}, Pascal~\cite{de2021part}, PACO~\cite{ramanathan2023paco} (Figure~\ref{fig::m2f_dataset}); MHP-v2~\cite{li2017multiple} for human part parsing; and TextOCR~\cite{singh2021textocr} for text segmentation. 
Additionally, we expanded the model's capabilities by adding 50 expert queries each for semantic parts and visual text regions, alongside the original 200 entity queries. 
We assessed Mask2Former+'s performance on 1000 images from validation splits from 4 grounding benchmarks, RefCOCO+~\cite{yu2016modeling}, PhraseCut~\cite{wu2020phrasecut}, ReasonSeg~\cite{lai2023lisa}, and TextVQA-X~\cite{singh2019towards}.
We use the Any-IoU~\cite{kamath2021mdetr} metric for evaluation, i.e., for each ground truth mask, we extract the most overlapped mask proposals and compute the IoU, then take the average. 
As Table~\ref{tab:m2f_res} demonstrates, Mask2Former+ shows consistent improvements across all domains, particularly in those significantly divergent from COCO. This highlights its enhanced adaptability and precision in a broader range of segmentation challenges, providing a good mask proposal model for \model.
We refer to Appendix~\ref{subsec:mask2former_data} for more details of the model and data.
 
\begin{table}[t!]
    \centering
    \resizebox{\columnwidth}{!}{
    \begin{tabular}{lcccc}
    \toprule
    \textbf{Model} & \textbf{RefCOCO+} & \textbf{PhraseCut} & \textbf{ReasonSeg} & \textbf{TextVQA-X} \\
    \cmidrule(r){1-1} \cmidrule(r){2-5}
    Mask2Former	            & 0.867 & 0.563	& 0.602 & 0.137	\\    
    Mask2Former+            & \textbf{0.873} & \textbf{0.624} & \textbf{0.745} & \textbf{0.446} \\
    \bottomrule 
    \end{tabular}}
    \vspace{-5pt}
    \caption{The average Any-IoU of the proposals on each dataset. The vanilla Mask2Former is trained on the COCO-Panoptic dataset and our Mask2Former+ is trained on our combined dataset. Mask2Former+ obtains a consistent improvement in all scenarios, especially in non-COCO domains. \vspace{-10pt}}    
    \label{tab:m2f_res}
\end{table}

\section{Our Dataset: M3G2}
\label{sec:formatting}

\begin{table}[!t]
    \centering
    \scalebox{0.7}{
    \begin{tabular}{lclcccccccr}
    \toprule
    \rowcolor[HTML]{C9DAF8}
    \multicolumn{1}{c}{Task} &
      \multicolumn{1}{c}{Dataset} &
      \multicolumn{3}{c}{Gr. Ann.} &
      \multicolumn{5}{c}{Sem. Gran.} &
      \multicolumn{1}{c}{\# Pairs} \\
    \rowcolor[HTML]{D8D8D8}
     &
      \multicolumn{1}{c}{} &
      \multicolumn{1}{c}{M} &
      \multicolumn{1}{c}{B} &
      \multicolumn{1}{c}{Po} &
      \multicolumn{1}{c}{S} &
      \multicolumn{1}{c}{Th} &
      \multicolumn{1}{c}{Pa} &
      \multicolumn{1}{c}{G} &
      \multicolumn{1}{c}{Tx} &
      \multicolumn{1}{r}{Train} \\
    \cmidrule(r){1-1} \cmidrule(r){2-2} \cmidrule(r){3-5} \cmidrule(r){6-10} \cmidrule(r){11-11} 

    \multirow{2}{*}{GCAP} & PNG               & \cmark & \cmark &                       & \cmark & \cmark &                       & \cmark &                       & 132k                  \\
                             & Flickr30K-Entity  &                       & \cmark &                       & \cmark & \cmark & \cmark & \cmark &                       & 149k                   \\
     \cmidrule(r){1-2} \cmidrule(r){3-5} \cmidrule(r){6-10} \cmidrule(r){11-11}
    \multirow{10}{*}{RES}    & RefCOCO           & \cmark & \cmark &                       & \cmark &                       &                       &                       &                       & 113k                  \\
                             & RefCOCO+          & \cmark & \cmark &                       & \cmark &                       &                       &                       &                       & 112k                  \\
                             & RefCOCOg          & \cmark & \cmark &                       & \cmark &                       &                       &                       &                       & 80k                   \\
                             & RefCLEF           & \cmark & \cmark &                       & \cmark &                       &                       &                       &                       & 105k                  \\
                             & gRefCOCO          & \cmark & \cmark &                       & \cmark &                       &                       &                       &                       & 194k                  \\
                             & PhraseCut         & \cmark & \cmark &                       & \cmark & \cmark & \cmark & \cmark &                       & 85k                   \\
                             & D-Cube             & \cmark & \cmark &                       & \cmark &                       &                       & \cmark &                       & 10k                   \\
                             & ReasonSeg         & \cmark & \cmark &                       & \cmark & \cmark & \cmark & \cmark &                       & 1k                    \\
                             & RIO               & \cmark & \cmark &                       & \cmark &                       &                       & \cmark &                       & 28k                   \\
                             & SK-VG             &                       & \cmark &                       & \cmark &                       &                       &                       &                       & 23k                   \\
     \cmidrule(r){1-2} \cmidrule(r){3-5} \cmidrule(r){6-10} \cmidrule(r){11-11}
    \multirow{8}{*}{GVQA}     & VizWiz-G  & \cmark & \cmark &                       & \cmark & \cmark &                       &                       & \cmark & 6k                    \\
                             & TextVQA-X         & \cmark & \cmark &                       &                       &                       &                       &                       & \cmark & 15k                   \\
                             & GQA               &                       & \cmark &                       & \cmark & \cmark & \cmark & \cmark &                       & 302k                  \\
                             & VQS               &                       & \cmark &                       & \cmark &                       &                       &                       &                       & 20k                   \\
                             & Shikra-BinaryQA   &                       & \cmark &                       & \cmark & \cmark & \cmark & \cmark &                       & 4k                    \\
                             & EntityCount       & \cmark & \cmark &                       & \cmark & \cmark & \cmark & \cmark &                       & 11k                   \\
                             & FoodSeg-QA        & \cmark & \cmark &                       & \cmark &                       &                       & \cmark &                       & 7k                    \\
                             & LVIS-QA           & \cmark & \cmark &                       & \cmark & \cmark &                       & \cmark &                       & 95k                   \\
     \cmidrule(r){1-2} \cmidrule(r){3-5} \cmidrule(r){6-10} \cmidrule(r){11-11}
    \multirow{16}{*}{RD}     & RefCOCO-REG       & \cmark & \cmark & \cmark & \cmark &                       &                       &                       &                       & 17k                   \\
                             & RefCOCO+-REG      & \cmark & \cmark & \cmark & \cmark &                       &                       &                       &                       & 17k                   \\
                             & RefCOCOg-REG      & \cmark & \cmark & \cmark & \cmark &                       &                       &                       &                       & 22k                   \\
                             & gRefCOCO-REG      & \cmark & \cmark & \cmark & \cmark &                       &                       &                       &                       & 20k                   \\
                             & VG-SpotCap        &                       & \cmark & \cmark & \cmark & \cmark & \cmark & \cmark &                       & 247k                  \\
                             & V7W               &                       & \cmark & \cmark & \cmark &                       &                       &                       &                       & 23k                   \\
                             & PointQA     &                       &                       & \cmark & \cmark &                       &                       &                       &                       & 64k                   \\
                             & VCR          &                       & \cmark & \cmark & \cmark &                       &                       &                       &                       & 156k                   \\
                             & ShikraRD          &                       & \cmark & \cmark & \cmark & \cmark & \cmark & \cmark &                       & 2k                    \\
                             & SVIT-RD           &                       & \cmark & \cmark & \cmark & \cmark & \cmark & \cmark &                       & 33k                   \\
                             & Guesswhat & \cmark & \cmark & \cmark & \cmark &                       &                       &                       &                       & 193k                   \\
                             & VG-RefMatch       &                       & \cmark & \cmark & \cmark & \cmark & \cmark & \cmark &                       & 247k                  \\
                             & HierText          & \cmark & \cmark & \cmark &                       &                       &                       &                       & \cmark & 6k                    \\

    \midrule
    \rowcolor[HTML]{D8D8D8}
    \multicolumn{10}{l}{M3G2 (Total)} &
      2.5M \\
      \bottomrule
    \end{tabular}}
    \vspace{-5pt}
    \caption{Summary of datasets included in M3G2. The datasets are grouped by four task types: \underline{G}rounded \underline{I}mage \underline{C}aptioning, \underline{R}eferring \underline{E}xpression \underline{S}egmentation, \underline{G}rounded \underline{V}isual \underline{Q}uestion \underline{A}nswering, and \underline{R}eferential \underline{D}ialogue. We show the availability of \underline{Gr}ounding \underline{Ann}otations  (\underline{B}ox, \underline{M}ask, and \underline{Po}inter inputs), the \underline{Sem}antic \underline{Gran}ularity (\underline{S}tuff, \underline{Th}ings, \underline{Pa}rts, \underline{G}roups, and \underline{T}e\underline{x}t), and the number of text-image pairs for training. \vspace{-20pt}}
    \label{tab:dataset_short}
\end{table}

In this section, we introduce M3G2, a \textbf{M}ulti-\textbf{M}odal \textbf{M}ulti-\textbf{G}rained \textbf{G}rounding dataset consisting of 2.5M text-image pairs for visually grounded instruction tuning, consisting of 36 sub-problems derived and augmented from 27 existing datasets.
We re-organize and augment public datasets of language grounding, visual question answering, referring expression segmentation, and referring expression generation into various forms of visually grounded dialogue for grounded instruction tuning, outlined briefly in Table~\ref{tab:dataset_short}.
The dataset is categorized into four main types: (1) Grounded Image Captioning (GIC), (2) Referential Expression Segmentation (RES), (3) Grounded Visual Question Answering (GVQA), and (4) Referential Dialog (RD). 
We provide illustrated descriptions of our prompt design, accompanied by examples of each task type as depicted in Figure~\ref{fig::dataset}.
We detail the task schema in the following sections and provide the complete sets of templates in Appendix~\ref{app:dataset}.

\subsection{Grounded Image Captioning (GIC)}

The task of \textit{grounded image captioning} requires the model to produce a narrative for the visual scene, and accurately identify and associate the groundable phrases with their respective binary segmentation masks. 
The objective of this task is to empower the model to articulate the scene while acknowledging various visual elements and their spatial interrelations.
We incorporate the Panoptic Narrative Grounding (PNG) dataset~\cite{kirillov2019panoptic} for dense and detailed scene descriptions, as well as the Flickr30K-Entity dataset~\cite{plummer2015flickr30k} for concise descriptions of the salient contents in the image. 
We create a collection of task prompt templates that instruct the model to describe the image either in detail or briefly.

\vspace{-2pt}
\subsection{Referring Expression Segmentation (RES)}
\vspace{-2pt}

In contrast to previous tasks, the \textit{referring expression segmentation} task requires that the model generates a segmentation mask based on a given referring expression. 
Besides the RefCOCO series~\cite{yu2016modeling,mao2016generation,liu2023gres}, we have further leveraged existing RES benchmarks~\cite{rohrbach2016refclef,wu2020phrasecut,xie2023described,wu2023advancing,lai2023lisa} for this purpose.
To frame a RES task, our prompts clearly direct the model to focus on the segmentation aspect of the given referring expression. 
An example prompt in Figure~\ref{fig::dataset} could be \texttt{Localize the sandwich on the plate at the far end}. 
The model's expected output would repeat the provided referring expression and accompany it with the correct segmentation mask, as in \texttt{Here it is: <GRD> the sandwich on the plate at the far end </GRD>} with a correct mask associated.

\subsection{Grounded Visual Question Answering (GVQA)}

\begin{figure}[!t]
    \centering
    \includegraphics[width=1.0\linewidth]{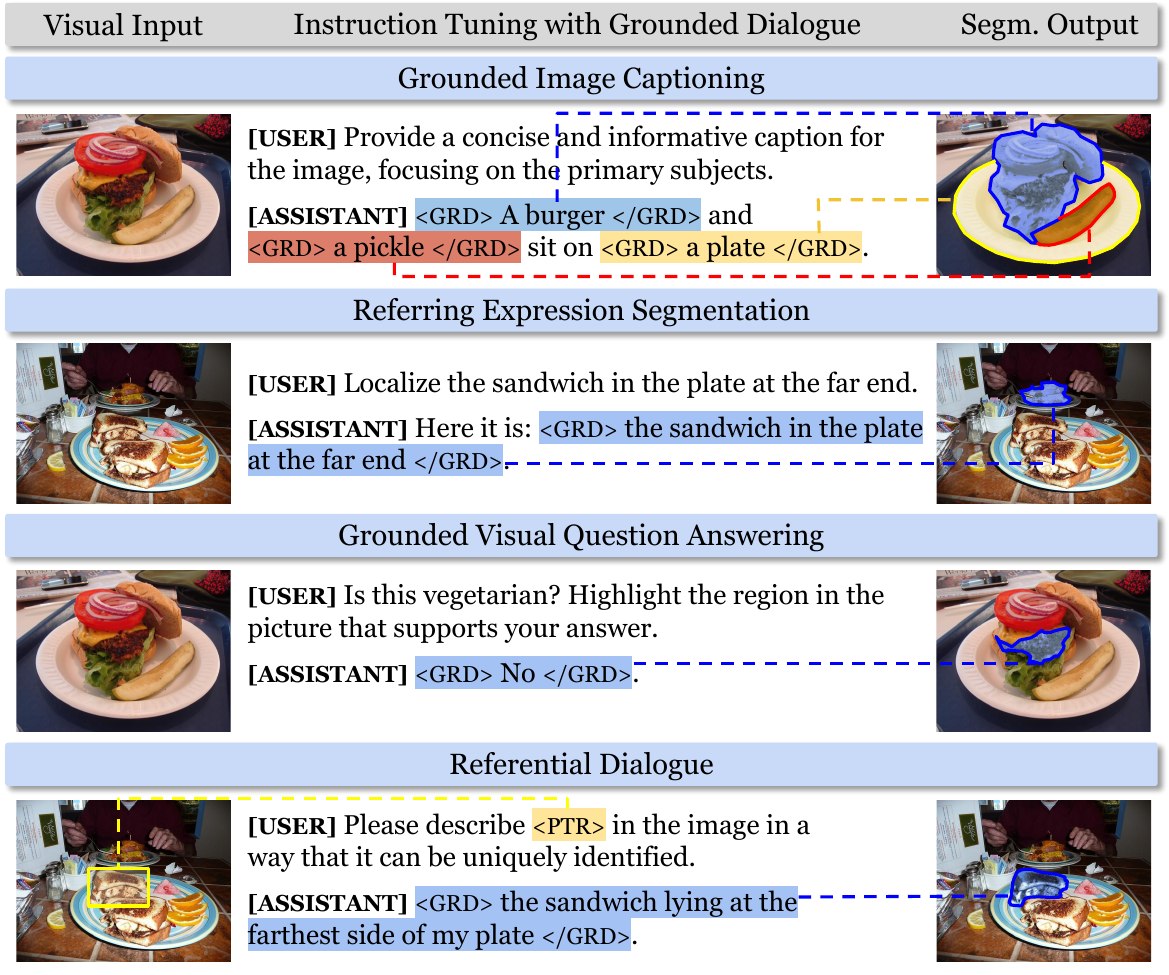}
    \vspace{-17pt}
    \caption{The M3G2 dataset for grounded visual instruction tuning. M3G2 is a diverse dataset of multiple granularities, unifying 4 different task types with visually grounded dialogue. \vspace{-15pt}}
    \label{fig::dataset}
\end{figure}

The task of \textit{grounded visual question answering} requires the model to comprehend a question (with optional pointers) and to produce an answer that is grounded to a binary segmentation mask that justifies the answer.
The goal of this task is to enable natural QA-based interaction with users and reasoning in the model with grounded explanations.
Specifically, we harvest and adapt a collection of public VQA datasets with grounding annotations~\cite{chen2022vizwizg,hudson2019gqa,gan2017vqs,chen2023shikra}, QA benchmarks on visual text~\cite{rao2021first}, and create templated QA pairs from segmentation datasets~\cite{gupta2019lvis,qi2023high}.
Our prompt templates instruct the model to respond either as open-ended answers or by selecting from multiple choices, with the response anchored to a segmentation mask (Figure~\ref{fig::dataset}).

\vspace{-2pt}
\subsection{Referential Dialogue (RD)}
\vspace{-2pt}

The task of \textit{referential dialogue} requires the model to conduct dialogue communication with users, especially when conditioned on user-provided spatial prompts.
This includes existing RD datasets~\cite{zhu2016visual7w,mani2020point,zellers2019recognition,chen2023shikra,zhao2023svit}, multi-turn augmentations from segmentation datasets~\cite{krishna2017visual,de2017guesswhat,long2022towards} as well as the \textit{referring expression generation (REG)} task the RefCOCO series~\cite{yu2016modeling,mao2016generation,liu2023gres}.
The REG task differs from the region captioning task in that it demands the description to be a referring expression that distinctly identifies the targeted object. 
Effective REG calls for the model to engage in dialogue interactions cooperatively, adhering to the Gricean Maxims~\cite{grice1975logic} which dictate that communication should be as informative, truthful, relevant, and clear as necessary.

\section{Experiment and Analysis}

\begin{table*}
    \centering
    \scalebox{0.81}{
    \begin{tabular}{lcccccccccccccc}
    \toprule
    & \multicolumn{8}{c}{\textbf{Single Instance}} & \multicolumn{2}{c}{\textbf{Multi-/No Instance}} & \multicolumn{3}{c}{\textbf{Reasoning}} \\
    \cmidrule(r){2-9} \cmidrule(r){10-11} \cmidrule(r){12-14}
    \multicolumn{1}{c}{\textbf{Model}} & \multicolumn{3}{c}{\textbf{RefCOCO}} & \multicolumn{3}{c}{\textbf{RefCOCO+}} & \multicolumn{2}{c}{\textbf{RefCOCOg}} & \textbf{gRefCOCO} & \textbf{PhraseCut} & \textbf{ReasonSeg} & \multicolumn{2}{c}{\textbf{RIO}} \\
    \cmidrule(r){2-4} \cmidrule(r){5-7}  \cmidrule(r){8-9}  \cmidrule(r){10-10} \cmidrule(r){11-11} \cmidrule(r){12-12} \cmidrule(r){13-14}
    & \textbf{val} & \textbf{test-A} & \textbf{test-B} & \textbf{val} & \textbf{test-A} & \textbf{test-B} & \textbf{val-u} & \textbf{test-u} & \textbf{val} & \textbf{test}  & \textbf{val}  & \textbf{test-c} & \textbf{test-u}\\
    \midrule
    \rowcolor[HTML]{C9DAF8}
    \textit{Specialist} & & & & & & & & &  &  &  &  &\\
    MDETR~\cite{kamath2021mdetr}    & -    & -    &  -   &  - &  - &  - &  - &  - &  - & 53.7 & - & 44.1 & 22.0 \\
    CRIS~\cite{wang2022cris}        & 70.5 & 73.2 & 66.1 & 62.3 & 68.1 & 53.7 & 59.9 & 60.4 & 55.3 & - & - & - & - \\
    LAVT~\cite{yang2022lavt}        & 72.7 & 75.8 & 68.8 & 62.1 & 68.4 & 55.1 & 61.2 & 62.1 & 58.4 & - & - & - & - \\
    ReLA~\cite{liu2023gres}        & 73.8 & 76.5 & 70.2 & 66.0 & 71.0 & 57.7 & 65.0 & 66.0 & 63.6 & - & 22.4 & - & - \\
    PolyFormer~\cite{liu2023polyformer}  & 76.0 & 78.3 & 73.3 & 69.3 & 74.6 & 61.9 & 69.2 & 70.2 & - & - & - & 48.8 & 26.8 \\ 
    UNINEXT-H~\cite{Yan2023uninext}   & 82.2 & 83.4 & 81.3 & 72.5 & 76.4 & 66.2 & 74.7 & 76.4 & - & - & - & - & - \\
    \midrule
    \rowcolor[HTML]{C9DAF8}
    \textit{Generalist} & & & & & & & & &  &  &  & & \\
    
    LISA$_{\textrm{7B}}$~\cite{lai2023lisa}       & 74.1 & 76.5 & 71.1 & 62.4 & 67.4 & 56.5 & 66.4 & 68.5 & - & - & 44.0 & - & -\\
    LISA$_{\textrm{7B}}$ (FT)~\cite{lai2023lisa}    & 74.9 & 79.1 & 72.3 & 65.1 & 70.8 & 58.1 & 67.9 & 70.6 & - & - & 52.9 & - & -\\
    \icon$_{\textrm{7B}}$       & 78.5 & 79.9 & 75.7 & 70.5 & 75.0 & 64.9 & 74.1 & 74.6 & 66.7 &  54.5 & 56.2 & 57.9 & 33.9 \\
    \bottomrule 
    \end{tabular}}
    \vspace{-8pt}
    \caption{Results on 7 Referring Expression Segmentation (RES) benchmarks with single instance queries~\cite{kazemzadeh2014referitgame,mao2016refcocog}, multi-/null instance queries \cite{liu2023gres,wu2020phrasecut} and reasoning-based queries \cite{lai2023lisa,qu2023rio}. We report cIoU for RefCOCO/+/g and mIoU for other benchmarks, respectively. \vspace{-5pt}}
    \label{tab:res}
\end{table*}
\begin{table*}
    \centering
    \begin{minipage}[t]{0.6\columnwidth}
    \centering
        \scalebox{1.05}{
            \begin{tabular}{lcc}
            \toprule
             & \multicolumn{2}{c}{\textbf{Flickr30K-E}} \\            
            \textbf{Model} & \textbf{R@1\textsubscript{val}}  & \textbf{R@1\textsubscript{test}} \\
            \midrule
            Shikra$_{\textrm{13B}}$ & 77.4 & 78.4\\
            Ferret$_{\textrm{13B}}$ & 81.1 & 84.8\\
            \midrule
            Shikra$_{\textrm{7B}}$ & 75.8 & 76.5\\
            Ferret$_{\textrm{7B}}$ & 80.4 & 82.2\\
            \icon$_{\textrm{7B}}$ & 79.2 & 79.8\\
            \bottomrule 
            \end{tabular}
        }
        \vspace{-8pt}
        \caption{Top-1 box recall results on Flickr30K-Entity~\cite{plummer2015flickr30k}.\vspace*{-15pt}}
        \label{tab:flickr-ground}
    \end{minipage}
    \hfill
    \raisebox{1.0\height}{
    \hspace*{-20pt}
    \begin{minipage}[t]{0.75\columnwidth}
    \centering
    \begin{minipage}[t]{1.0\columnwidth}
        \scalebox{0.85}{
            \begin{tabular}{lccccccc}
            \toprule
             & \multicolumn{5}{c}{\textbf{{PNG}}} \\    
            \textbf{Model} & \textbf{AR} & \textbf{AR\textsubscript{th}} & \textbf{AR\textsubscript{st}} & \textbf{AR\textsubscript{s}} & \textbf{AR\textsubscript{p}}  \\
            \cmidrule(r){1-1} \cmidrule(r){2-6}
            PiGLET	& 65.9 & 	64.0	& 68.6	& 67.2	&  54.5  \\    
            \icon$_{\textrm{7B}}$ & 66.8 & 65.0 & 69.4 & 70.4 & 57.7 \\
            \bottomrule 
            \end{tabular}
        }
        \vspace{-10pt}
        \caption{Phrase grounding results on PNG~\cite{gonzalez2021panoptic}.}
        \label{tab:png}
    \end{minipage}
    \hfill
    \begin{minipage}[t]{1.0\columnwidth}
        \scalebox{0.875}{
            \begin{tabular}{lccccccc}
            \toprule
             \textbf{Model} & & & \textbf{TextVQA-X [mIoU]} & & \\  
            \midrule
            SAB	& & & 29.0  & &  \\    
            \icon$_{\textrm{7B}}$ & & & 39.8 & & \\
            \bottomrule 
            \end{tabular}
        }
        \vspace{-8pt}
        \caption{Visual text QA results on the TextVQA-X~\cite{singh2019towards} validation set. \vspace{-5pt}}
        \label{tab:textvqa}
    \end{minipage}
    \end{minipage}}
    \hfill
    \begin{minipage}[t]{0.6\columnwidth}
    \centering
        \scalebox{1.05}{
        \hspace*{-25pt}
        \begin{tabular}{lcc}
            \toprule
             &  \textbf{PointQA$_{\textrm{Twice}}$} & \textbf{V7W}  \\
            \multirow{1}{*}{\textbf{Model}} & \textbf{Acc} & \textbf{Acc} \\
            \cmidrule(r){1-1} \cmidrule(r){2-3}
            Shikra$_{\textrm{13B}}$	& 70.3 & 85.3		\\ 
            \small{GPT4RoI}$_{\textrm{13B}}$  & -  & 84.8   \\  
            \midrule
            Shikra$_{\textrm{7B}}$	& - & -		\\ 
            \small{GPT4RoI}$_{\textrm{7B}}$  & -  & 81.8   \\        
            \icon$_{\textrm{7B}}$ & 72.4 & 85.5 \\
            \bottomrule 
            \end{tabular}
        }
        \vspace{-8pt}
        \caption{Results on PointQA$_{\textrm{Twice}}$~\cite{mani2020point} and V7W~\cite{zhu2016visual7w} test sets. \vspace*{-15pt}}
        \label{tab:ptrvqa_c}
    \end{minipage}
    \vspace{-10pt}
\end{table*}

\subsection{Implementation}

\paragraph{Learning from Both Box and Mask Supervision.} 
In the M3G2 dataset, not all sub-datasets include mask supervision.
We employ different loss functions to effectively benefit from grounded supervision from both mask and box annotations. 
When the mask annotations are available, we apply the dice loss $\mathcal{L}_{\text{dice}}$ and binary cross-entropy loss $\mathcal{L}_{\text{bce}}$ between the predicted grounding masks and the ground truth masks of each phrase, following~\citet{cheng2022masked}. 
When the box annotations are present, we apply the projection loss $\mathcal{L}_{\text{proj}}$ as introduced by \citet{tian2021boxinst}.
The final loss calculation is a linear combination of the language modeling loss $\mathcal{L}_{\text{lm}}$ and these mask-related losses.
We refer to Appendix~\ref{app:implement} for more details and explanations of these loss terms. 

\vspace*{-10pt}
\paragraph{Parameter-Efficient Training Details.} 
We adopt the LLaMA2-7B model ~\cite{touvron2023llama2} as our base LLM, and initialized the weight from LLaVA-1.5~\cite{liu2023improved}. For the vision encoders, we use the OpenAI CLIP@336~\cite{radford2021clip} model and DINOv2-L/14-reg~\cite{darcet2023vision} pretrained checkpoints.
We freeze all the parameters of Mask2Former+, CLIP, and DINOv2 during training. We use Low-Rank Adaptation (LoRA)~\cite{hu2021lora} with $r=16$ and $\alpha=16$ to tune the LLM, including all the linear layers, input embeddings, and the LM head.
We train all the new components introduced for connecting these models, including the MLP projection layer of CLIP and DINOv2, and the mask retrieval head. 
As a result, less than 2\% of the total parameters are trainable in the whole model.
We use the AdamW optimizer~\cite{loshchilov2017decoupled} with an initial learning rate of 2e-4 and a cosine annealing rate. We train our model on the balanced sampled M3G2 dataset for 2 epochs, which takes around 2 days using 8 40G A100 GPUs.
\vspace{-2pt}
\subsection{Generalist in Grounded Vision-Language Tasks}
\vspace{-5pt}

We first demonstrate \model's capabilities as a generalist model for three different types of grounded vision-language tasks. 
It's worth noting that, unlike previous work that needs dataset-specific fine-tuning on each of the tasks, \model~can achieve comparable performance on all the tasks directly after training on M3G2, i.e., all the reported results from our model are from a single set of weights without any dataset-specific fine-tuning.



\begin{figure*}[!t]
    \centering
    \begin{subfigure}[t]{.48\textwidth}
        \centering
        \includegraphics[width=1.0\linewidth]{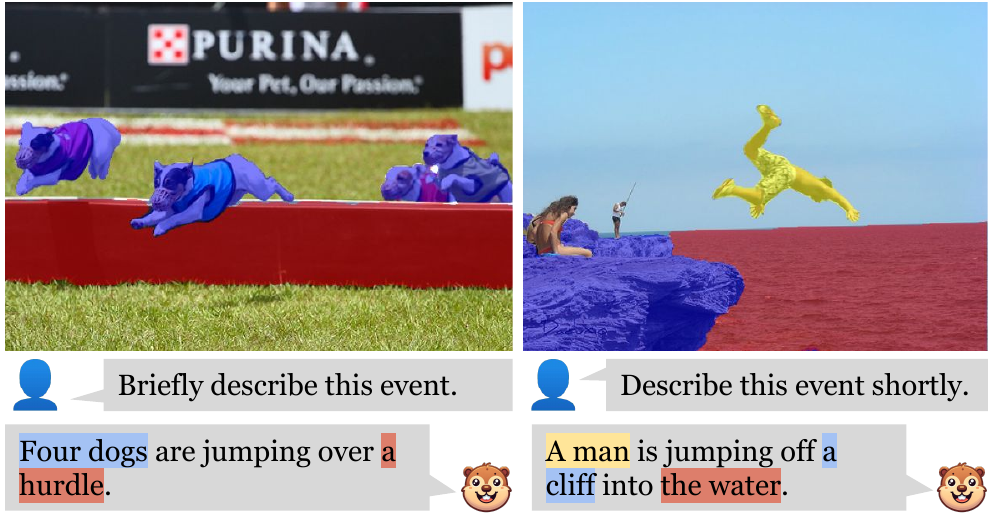}
    \caption{Grounded short caption generation on Flickr30K-Entity. While only box supervisions are available for this dataset, \model~generalize to pixel-level grounding after joint training on M3G2.}
    \label{fig:case_caption_1}
    \end{subfigure}
    ~
    \begin{subfigure}[t]{0.49\textwidth}
        \centering
        \includegraphics[width=1.0\linewidth]{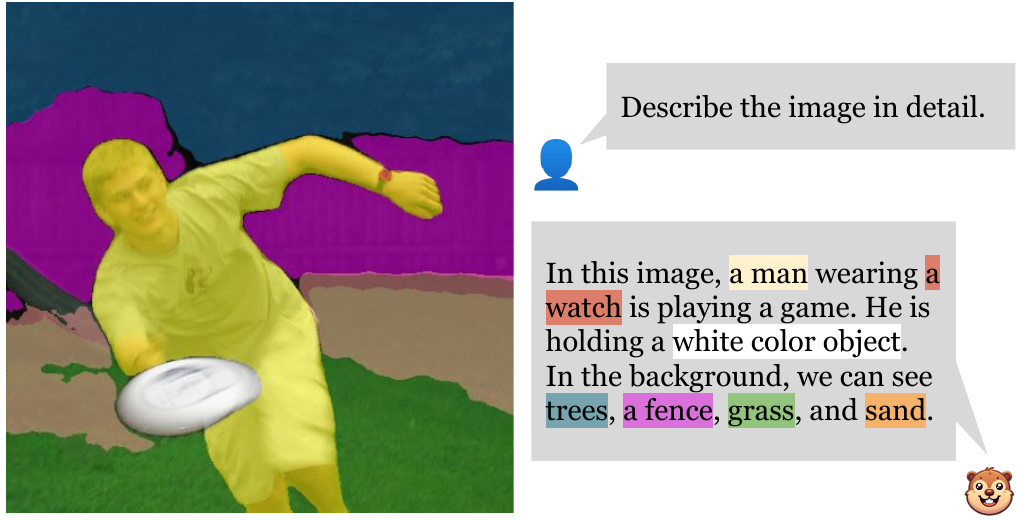}
    \caption{Grounded detailed narrative generation on PNG. \model~successfully generalize to grounding a novel category \textit{watch} in the generated caption, which is not included in the 80 categories of PNG annotation. \vspace{-5pt}}
    \label{fig:case_caption_2}
    \end{subfigure}
    \vspace{-5pt}
    \caption{Examples of \model's performance in grounded image captioning. \vspace{-5pt}}
    \label{fig:demo-2}
\end{figure*}

\vspace{-10pt}
\paragraph{Language Grounding To Segmentation.}
We start by evaluating the model on language grounding tasks, which takes text as input and generates segmentation masks as output.
We assess \model~on Referential Expression Segmentation (RES)~\cite{kazemzadeh2014referitgame} and Caption Phrase Grounding (CPG) tasks. 
While traditional RES benchmarks~\cite{kazemzadeh2014referitgame,mao2016refcocog} focus on single-instance referents requiring primarily visual understanding, we expanded our evaluation to include complex scenarios involving multi-instance or negative queries \cite{wu2020phrasecut,liu2023gres}, and those necessitating common sense reasoning~\cite{lai2023lisa,qu2023rio}. 
For single-instance RES, we report the cIoU; and for the other benchmarks, we report the mIoU.
The results, as detailed in Table~\ref{tab:res}, show \model~outperforming the generalist model LISA across all benchmarks and achieving significant improvements over specialist models in multi-instance, null, and reasoning-based RES tasks. 
It also performs comparably on the competitive RefCOCO series. 
For CPG tasks, which involve grounding all phrases in a caption and demand a deep understanding of the context for coreference resolution, we first evaluated \model~on the Flickr30K-Entity dataset~\cite{plummer2015flickr30k}. 
Since this dataset only has box annotations, we convert the mask predictions of our model to box and compute the top-1 box recall following the merged-box protocol (All-IoU)~\cite{kamath2021mdetr}. 
Despite not specializing in predicting boxes, \model~still outperforms Shikra 7B/13B~\cite{chen2023shikra} and is on par with Ferret-7B~\cite{you2023ferret} in a concurrent work (Table~\ref{tab:flickr-ground}). 
Additionally, on the PNG dataset \cite{kirillov2019panoptic} which tests phrase grounding in longer narratives, \model~surpasses the previous state-of-the-art model, PiGLET~\cite{gonzalez2023piglet}, in all metrics including average recall of grounding masks and detailed scores for things, stuffs, and singular and plural entities (Table~\ref{tab:png}).

\begin{table}
    \centering
    \resizebox{\columnwidth}{!}{
    \begin{tabular}{lccccc}
    \toprule
    \textbf{Model} & \textbf{Bleu-4} & \textbf{METEOR} & \textbf{CIDEr} & \textbf{SPICE} & \textbf{F1\textsubscript{all}} \\
    \midrule
    Shikra$_{\textrm{13B}}$ & - & - & 73.9 & - & - \\
    Ferret$_{\textrm{13B}}$ & 37.0  & 25.5  & 76.1 & 18.3 & 15.1 \\
    \midrule
    Ferret$_{\textrm{7B}}$ & 35.1  & 24.6  & 74.8 & 18.0 & 15.0 \\
    \icon$_{\textrm{7B}}$ & 36.7 & 26.5 & 91.3 & 20.4 & 32.1 \\
    \bottomrule 
    \end{tabular}}
    \vspace*{-5pt}
    \caption{Grounded Captioning on Flickr30K-Entity~\cite{plummer2015flickr30k}. \vspace*{-5pt}}
    \label{tab:flickr_cap}
\end{table}

\vspace{-10pt}
\paragraph{Grounded Language Generation.}

Our model excels in generating language that accurately grounds to segmentation masks during user conversations. 
Quantitatively, we assess grounded captioning on the Flickr30K-Entity dataset~\cite{plummer2015flickr30k}, employing standard text generation metrics such as Bleu-4~\cite{papineni2002bleu}, METEOR~\cite{banerjee2005meteor}, CIDEr~\cite{vedantam2015cider}, and SPICE~\cite{anderson2016spice} for language quality; and the F1\textsubscript{all} score for grounding accuracy following~\citet{you2023ferret}. 
As shown in Table~\ref{tab:flickr_cap}, \model~significantly surpasses existing box-based grounded MLLMs, even their 13B versions, in both language quality and grounding accuracy. 
This improvement is hypothesized to stem from the diverse task distribution in our M3G2 dataset. 
We show some generated captions in Figure~\ref{fig:demo-2}, with a highlight of box-to-pixel generalization (Figure \ref{fig:case_caption_1}) and novel category grounding (Figure \ref{fig:case_caption_2}). See the Appendix for more examples. 
For groundable question answering, we evaluate on the TextVQA-X benchmark~\cite{rao2021first}.
\model~outperforms the state-of-the-art specialist model SAB~\cite{khoshsirat2023sentence} by a significant margin, as measured by the mean IoU of the predicted mask (Table~\ref{tab:textvqa}).

\begin{figure}[!t]
    \centering
    \includegraphics[width=1.0\linewidth]{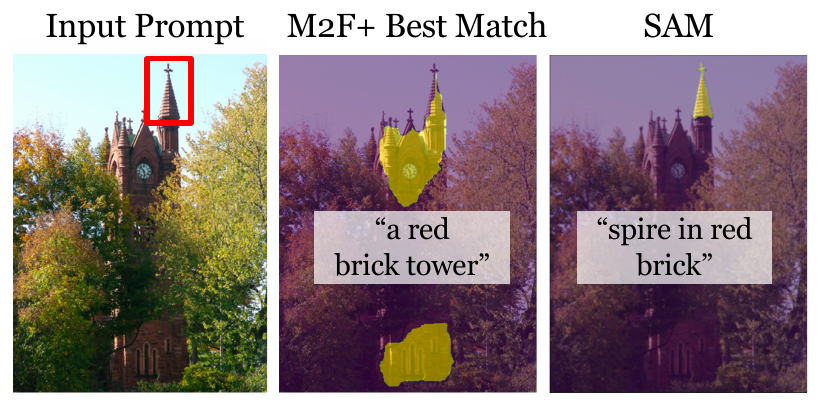}
    \vspace{-10pt}
    \caption{Region caption using the best match proposal from Mask2Former+ versus from SAM. Mask2Former+ fails to propose the exact mask of the spire, leading to a less precise caption. \vspace{-10pt}}
    \label{fig:pointer-example}
\end{figure}

\vspace{-12pt}
\paragraph{Spatial Prompt Understanding.}
For grounded MLLMs, accurately interpreting multimodal instructions is essential, particularly in interactive tasks. 
We evaluated its performance on two pointer-based QA benchmarks, PointerQA$_{\textrm{Twice}}$~\cite{mani2020point} and V7W~\cite{zhu2016visual7w}, which require the model to answer questions guided by spatial prompts, such as bounding boxes. 
The model is tasked to generate free-form textual answers in PointerQA$_{\textrm{Twice}}$, and selects from multiple-choice options in V7W. 
\model~demonstrates superior performance in these benchmarks, outperforming previous models as shown in Table~\ref{tab:ptrvqa_c}. 
This highlights its effectiveness in spatial understanding and response accuracy. 
To further demonstrate the effectiveness of using SAM for the pointer-to-mask conversion, we show the best-matched mask proposal from our Mask2Former+ model in comparison to the mask from SAM in Figure~\ref{fig:pointer-example}. 
While the best match proposal from the Mask2Former+ model includes a broader area, the SAM-generated mask offers a more precise representation of the specified region, potentially leading to a more accurate caption. 


\begin{table}
    \centering
    \scalebox{0.75}{
    \begin{tabular}{lccccc}
    \toprule
    Model            & Accuracy & Precision & Recall & F1 Score & Yes (\%) \\ 
    \midrule
    \rowcolor[HTML]{C9DAF8}
    \multicolumn{6}{l}{\textit{Random}}                                                     \\
    mPLUG-Owl        & 53.30    & 51.71     & 99.53  & 68.06    & 96.23    \\
     LLaVA            & 54.43    & 52.32     & 99.80  & 68.65    & 95.37    \\
     MultiModal-GPT   & 50.03    & 50.02     & 100.00 & 66.68    & 99.97    \\
     MiniGPT-4        & 77.83    & 75.38     & 82.67  & 78.86    & 54.83    \\
     InstructBLIP     & 88.73    & 85.08     & 93.93  & 89.29    & 55.20    \\
     Shikra-13B        & 86.90    & 94.40     & 79.26  & 86.19    & 43.26    \\
     Ferret-13B        & 90.24    & 97.72     & 83.00  & 89.76    & 43.26    \\
    \icon$_{\textrm{7B}}$         & 91.03    & 85.80     & 96.40  & 90.79    & 45.88    \\
    \midrule
    \rowcolor[HTML]{C9DAF8}
    \multicolumn{6}{l}{\textit{Popular}}                                                           \\
     mPLUG-Owl        & 50.63    & 50.32     & 99.27  & 66.79    & 98.63    \\
     LLaVA            & 52.43    & 51.25     & 99.80  & 67.72    & 97.37    \\
     MultiModal-GPT   & 50.00    & 50.00     & 100.00 & 66.67    & 100.00   \\
     MiniGPT-4        & 68.30    & 64.27     & 82.40  & 72.21    & 64.10    \\
     InstructBLIP     & 81.37    & 75.07     & 93.93  & 83.45    & 62.57    \\
     Shikra-13B        & 83.97    & 87.55     & 79.20  & 83.16    & 45.23    \\
     Ferret-13B        & 84.90    & 88.24     & 80.53  & 84.21    & 45.63    \\
    \icon$_{\textrm{7B}}$         & 90.13    & 85.93     & 93.81  & 89.70    & 45.80    \\
    \midrule
    \rowcolor[HTML]{C9DAF8}
    \multicolumn{6}{l}{\textit{Adversarial}}                                                      \\
     mPLUG-Owl        & 50.67    & 50.34     & 99.33  & 66.82    & 98.67    \\
     LLaVA            & 50.77    & 50.39     & 99.87  & 66.98    & 99.10    \\
     MultiModal-GPT   & 50.00    & 50.00     & 100.00 & 66.67    & 100.00   \\
     MiniGPT-4        & 66.60    & 62.45     & 83.27  & 71.37    & 66.67    \\
     InstructBLIP     & 74.37    & 67.67     & 93.33  & 78.45    & 68.97    \\ 
     Shikra-13B        & 83.10    & 85.60     & 79.60  & 82.49    & 46.50    \\
     Ferret-13B        & 82.36    & 83.60     & 80.53  & 82.00    & 48.18    \\
    \icon$_{\textrm{7B}}$         & 86.33    & 85.93     & 86.63  & 86.28    & 49.60 \\
    \bottomrule
    \end{tabular}}
    \vspace{-10pt}
    \caption{Object hallucination results on the POPE~\cite{li2023evaluating} benchmark. \vspace{-20pt}}
    \label{tab:pope}
\end{table}

\vspace{-3pt}
\subsection{Trustworthiness and Transparency}

Beyond its superior performance as a grounding generalist, we highlight two key improvements for creating a more trustworthy and transparent agent. 

\vspace{-10pt}
\paragraph{Reduced Object Hallucination.}

Thanks to the varied task distribution and the inclusion of negative question-answering samples in M3G2 dataset, \model~significantly reduces object hallucination. 
We assessed this using the POPE~\cite{li2023evaluating} benchmark, which includes binary questions about object existence across three splits, each with a different object distribution (with an order of difficulty \textit{Random} $<$ \textit{Popular} $<$ \textit{Adversarial}). 
Remarkably, \model~consistently outperforms other models in both accuracy and F1 score across all splits, particularly on the more challenging ones. It shows an absolute improvement of 5.2\% in accuracy for \textit{Popular} and 4.0\% for \textit{Adversarial} over the previously best-performing model. This suggests that our model's enhanced grounding capability plays a significant role in mitigating the object hallucination problem.

\begin{figure}[!t]
    \centering
    \includegraphics[width=1.0\linewidth]{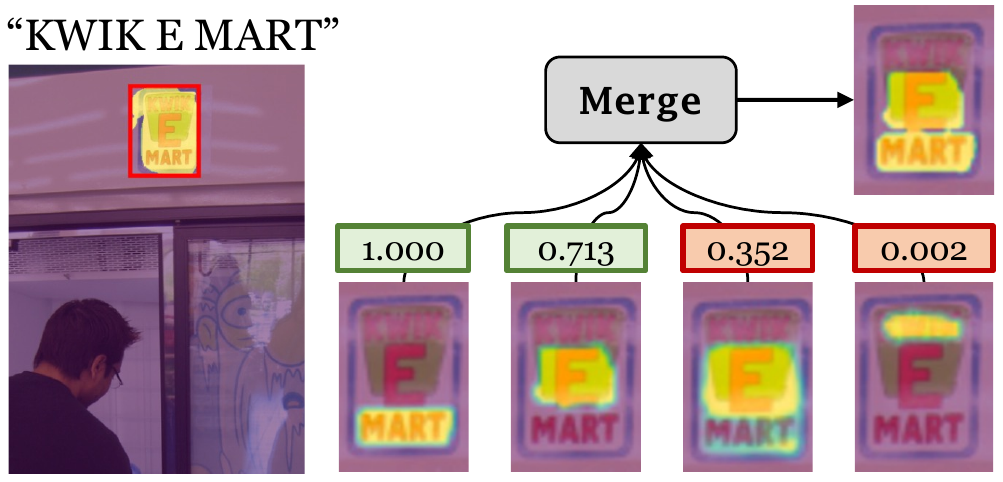}
    \vspace{-10pt}
    \caption{Illustration of a partially correct grounding. The grounding phrase and the ground truth mask are shown on the left. The top-4 mask proposals are presented, with highly-scored masks (green) selected for the merged mask, and low-scored masks (red) excluded. This illustrates the failure to recognize the word ``KWIK'' by the MLLM, despite its successful proposal. \vspace{-5pt}}
    \label{fig:proposal_case}
\end{figure}

\vspace{-10pt}
\paragraph{Explainability and Diagnosability.}
Another important highlight of \model~is its enhancement of explainability through the decoupled design of entity proposal and selection, as outlined earlier in section~\ref{subsec:grounding}. 
This is exemplified in the case study illustrated in Figure~\ref{fig:proposal_case}, which illustrates the mask proposal scoring and selective merging process of our model. 
We show the top-4 masks, where the higher-score masks are labeled in green while the lower-score masks are labeled in red. 
Users can easily interpret that the failure is due to the incapability of MLLM to recognize the word ``KWIK'', despite it being successfully localized and proposed as an entity candidate.

\begin{table}
    \centering
    \resizebox{\columnwidth}{!}{
    \begin{tabular}{lccc}
    \toprule
    \multirow{1}{*}{\textbf{Setups}} & \textbf{RefCOCO+} & \textbf{Flickr30K} & \textbf{TextVQA-X} \\
    \midrule    
    \rowcolor[HTML]{C9DAF8}
    \multicolumn{1}{l}{\textit{Mask Proposal Models}}  & &  &   \\
    Mask2Former  & \textbf{67.1}  & 69.0 & 9.8  \\
    Mask2Former+ & 66.6  & \textbf{77.2} & \textbf{34.0} \\
    \midrule    
    \rowcolor[HTML]{C9DAF8}
    \multicolumn{1}{l}{\textit{Entity Features}} & &  & \\
    CLIP & 59.8 & 75.0 & 32.0 \\  
    DINOv2 & 62.3 &  76.3 &  28.4   \\ 
    CLIP+DINOv2 & \textbf{66.6}  & \textbf{77.2} & \textbf{34.0} \\ 
    \rowcolor[HTML]{C9DAF8}
    \multicolumn{1}{l}{\textit{Grounding Query}} & &  & \\
    \texttt{<GRD>} only & 64.4 & 67.5 & \textbf{34.2} \\  
    \texttt{</GRD>} only & 64.4 &  \textbf{77.2} &  33.5   \\ 
    Sum & \textbf{66.6}  & \textbf{77.2} & 34.0 \\
    \midrule
    \rowcolor[HTML]{C9DAF8}
    \multicolumn{1}{l}{\textit{Eval Input Resolution}} & & &  \\
    224--480 & 54.7 & 67.2 & 27.6 \\ 
    480--640 & 65.5 & 76.7 & 27.6 \\ 
    800--1024 & \textbf{66.6}  & \textbf{77.2} & \textbf{34.0} \\ 
    \bottomrule 
    \end{tabular}}
    \vspace*{-5pt}
    \caption{Ablation study on model design choices and evaluation setups. Models are trained on RefCOCO+, Flickr30K, TextVQA-X and tested on corresponding validation sets. \vspace*{-10pt}}
    \label{tab:ablation_feature}
\end{table}

\subsection{Ablation Studies}
We performed ablation studies to validate our design decisions, training, and evaluating a subset of the M3G2 dataset that includes RefCOCO+, Flickr30K, and TextVQA. 
These cover a range of visual entities from various image sources and granularities. 
We start by comparing our Mask2Former+ with the original Mask2Former for mask proposal effectiveness. 
As indicated in Table~\ref{tab:ablation_feature}, the original Mask2Former performs slightly better on RefCOCO, as it is developed specificity on COCO object categories. 
However, Mask2Former+ significantly surpasses the original in domains with non-COCO entities. 
Our second set of experiments examined the choice of visual entity features. 
Although using either CLIP or DINOv2 features alone shows advantages in specific datasets, their combination consistently yields the best results across all datasets. 
To obtain a robust grounding query representation, we experimented with using the output embedding of the \texttt{<GRD>} token, the \texttt{</GRD>} token, and their sum. We found that the latter approach achieves the best overall results. 
Finally, we demonstrate that our decoupling design of the mask proposal model and MLLM allows for training at a lower resolution (320px) to expedite grounding training, while scaling up the resolution during evaluation enhances performance.

\vspace{-10pt}
\section{Related Work}
\vspace{-5pt}
\subsection{Multimodal Large Language Models}
\vspace{-5pt}
Building on the recent advance of large language models (LLMs), there is an increasing effort in adapting pretrained large language models for multimodal tasks, such as understanding and interpreting visual information~\cite{tsimpoukelli2021multimodal,alayrac2022flamingo}.
More recently, visual instruction tuning has gained much interest due to its surprising performance with a modest amount of data and computing resources.
Various models have been developed, noticeably MiniGPT4\cite{zhu2023minigpt}
, LLaVA~\citep{liu2023visual,liu2023improved} and concurrent models~\cite{dai2023instructblip,gong2023multimodal,ye2023mplug,wang2023visionllm,li2023otter}. 
Despite their promising performances, MLLMs often produce objects that are not presented in the given images, a phenomenon referred to as the \textit{object hallucination} problem~\cite{kayhan2021hallucination,dai2023plausible,li2023evaluating}.

\vspace{-3pt}
\subsection{MLLM with Language Grounding}
\vspace{-4pt}

The ability to connect language to their corresponding visual elements in the physical world, known as \textit{grounding}~\cite{harnad1990symbol}, is crucial in everyday human communication about our shared surroundings. 
Grounding datasets have been shown to benefit vision-language pre-training, both in terms of object-level recognition~\cite{li2022grounded} and language learning~\cite{ma2023world}.
Recent works unify text and grounding regions into token sequences~\cite{yang2022unitab,lu2022unified,wang2022ofa} in casual language modeling. 
Based on such paradigm, researchers have developed a family of grounded MLLM, including GPT4ROI~\cite{zhang2023gpt4roi}, Kosmos-2~\citep{peng2023kosmos}, Shikra~\citep{chen2023shikra}, PVIT~\cite{chen2023position}, BuboGPT~\cite{zhao2023bubogpt}, Qwen-VL~\cite{bai2023qwen}, and Ferret~\cite{you2023ferret}.
Despite their promising performance, these models focus on object grounding to bounding box, which cannot handle pixel-level grounding across various semantic granularities. Furthermore, it lacks the diagnosability and explainability in failure cases. We introduce \model~to fill this gap.

\vspace{-5pt}
\subsection{Language-Guided Semantic Localization}
\vspace{-3pt}
The field of language-guided semantic localization has a long history in the vision-language research community, requiring that the model localize a given referring expression with bounding boxes or segmentation masks.
This task has evolved from early attempts to understand simple referring expressions within images, such as the well-known RefCOCO series~\cite{yu2016modeling,mao2016generation} and their generalized variant~\cite{liu2023gres} that takes no-target and multi-target into account.
The integration of advanced language reasoning from LLMs has enabled research to tackle even more nuanced reasoning tasks that involve complex language contexts~\cite{zang2023contextual,pi2023detgpt,lai2023lisa}.
Notably, LISA~\citep{lai2023lisa} formulates a reasoning segmentation task to bring language-informed reasoning into semantic segmentation, and contributes a powerful baseline. 
Our model builds on these developments, but is designed to be more universally applicable as a grounded MLLM.

\vspace{-5pt}
\section{Conclusion}
\vspace{-2pt}

In this study, we introduce \model, a novel framework designed to enable pixel-level explainable grounding in large language models, leveraging holistic segmentation. 
The system builds upon a pre-trained mask proposal network to provide pixel-level visual features for the large language models, allowing them to retrieve segmentation mask proposals that can be used for grounding. We also present M3G2, 
a dataset of 1.9M training text-image pairs with 36 sub-problems derived from 27 existing datasets for visually grounded instruction tuning, facilitating precise vision-language alignment at the pixel level. We show that after training on M3G2, \model~achieves superior performance on various grounding tasks. Through extensive case studies, we further show that \model~unlocks explainability and diagnosability, and demonstrates better grounding towards occluded objects, groups of multiple instances, amorphous background regions, semantic parts of objects, and objects with irregular shapes.

\vspace{-5pt}
\section*{Limitations And Future Work}
\vspace{-5pt}
This work, while exciting, has several limitations that we acknowledge and aim to address in future research. Firstly, the datasets utilized to develop M3G2 consist of a blend of existing academic datasets. The quality of annotations in these datasets varies significantly, and they often lack comprehensive coverage of concepts. To enhance training efficiency, applying data filtering methods could help reduce the size of the dataset without compromising its effectiveness. Additionally, expanding the vision-language grounding data to a web-scale could significantly improve the comprehensiveness of grounding learning.

Secondly, our current model is limited to processing only single images. Although the entity-centric approach we adopted could theoretically extend to other modalities like 3D or video, this potential has not yet been empirically validated. Testing and validating our model on datasets relevant to these modalities would be a valuable direction for future research. This step is crucial to understanding the model's effectiveness across different types of application scenarios and further improving its usefulness.

\section*{Acknowledgements}
This work was supported by Amazon and NSF IIS-1949634. We would like to thank the anonymous reviewers for their valuable comments and suggestions.


{
    \small
    \bibliographystyle{ieeenat_fullname}
    \bibliography{main}
}

\clearpage
\clearpage
\setcounter{page}{1}
\setcounter{section}{0}
\renewcommand*{\thesection}{\Alph{section}}
\section*{\LARGE Appendix}

\section{Mask2Former+ Implementation Details}
\label{subsec:mask2former_data}

Our enhancement of the original Mask2Former model focuses on broadening its segmentation capabilities beyond the 134 common object categories it currently handles, which include 80 things and 55 stuffs as defined in the COCO dataset. 
The primary goal is to enable the model to recognize an expanded range of object categories, as well as segmentation masks of various levels of granularities, such as semantic parts and visual text regions. 

\begin{figure}[!t]
    \centering
    \vspace{-5pt}
    \includegraphics[width=1.0\linewidth]{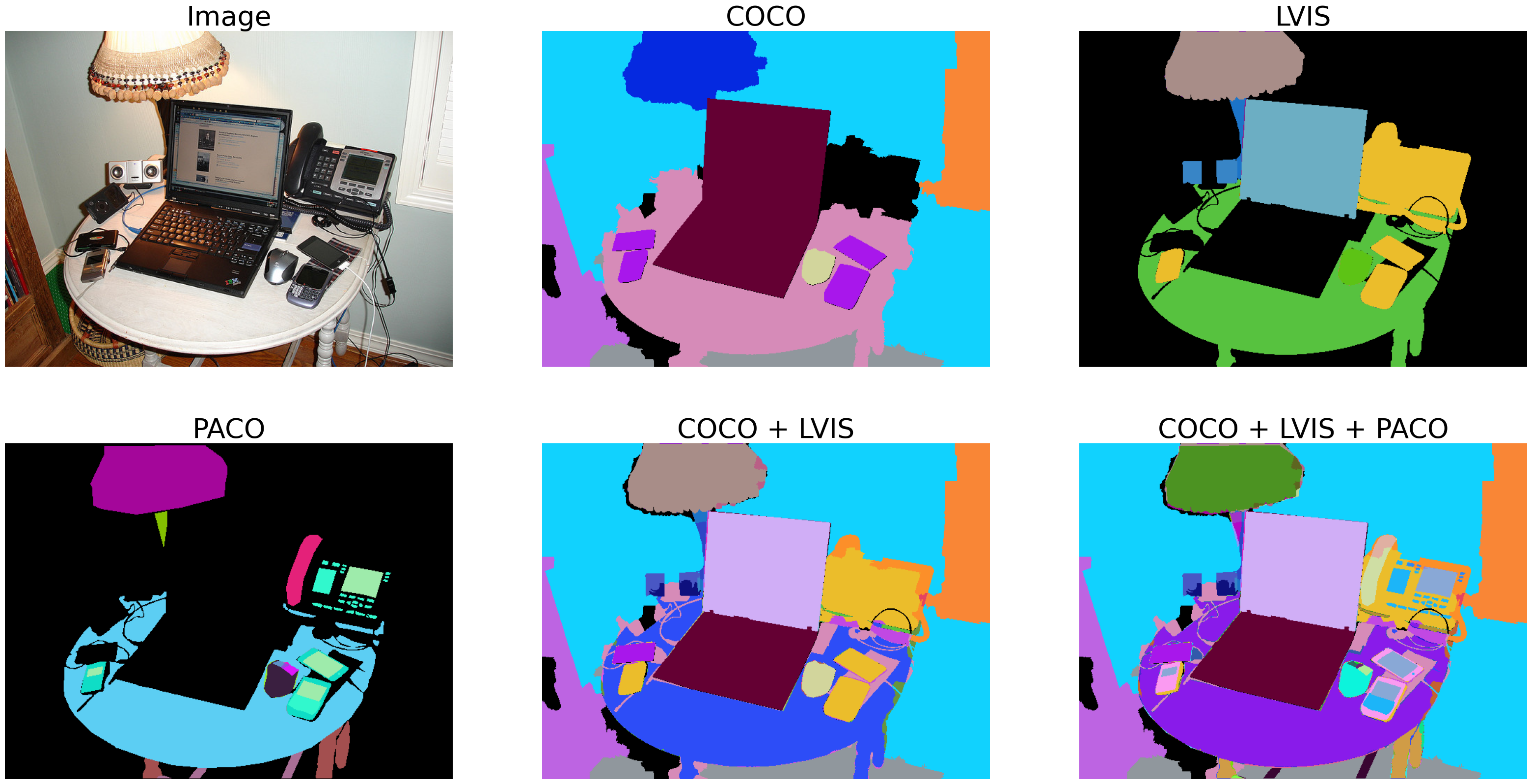}
    \vspace{-15pt}
    \caption{Illustrations of the merged segmentation annotations from COCO Panoptic, LVIS, and PACO datasets. \vspace{-15pt}}
    \label{fig::m2f_dataset}
\end{figure}

\paragraph{Training data.}
We have compiled a comprehensive dataset by combining multiple existing segmentation datasets. 
This ensemble encompasses a wide spectrum of entities (things and stuff), their semantic parts, and visual text, drawn from sources such as COCO \cite{caesar2018coco}, LVIS~\cite{gupta2019lvis}, Entity-v2~\cite{qi2023high}, Pascal~\cite{de2021part}, PACO~\cite{ramanathan2023paco}, MHP-v2~\cite{li2017multiple}, and TextOCR~\cite{singh2021textocr}. 
The resulting dataset comprises over 200K images and 4.5M masks, as summarized in Table~\ref{tab:mask2former}. 
Notably, the annotations from COCO, LVIS, and PACO are based on a shared set of COCO images.
We merged these annotations to ensure comprehensive mask proposal coverage, thereby providing holistic instance coverage within each image, as can be illustrated in  Figure~\ref{fig::m2f_dataset}.

\begin{table}[!h]
    \centering
    \vspace{-5pt}
    \scalebox{0.78}{
    \begin{tabular}{llcccrr}
    \toprule
    \rowcolor[HTML]{C9DAF8}
    \multicolumn{2}{c}{Dataset} & \multicolumn{3}{c}{Granularity}              & \multicolumn{2}{c}{Dataset Size} \\
    \rowcolor[HTML]{D8D8D8}
    Name                        & Split   & Entity               & Part                 & \multicolumn{1}{c}{Text}& \#Image & \#Masks  \\
    \cmidrule(r){1-2} \cmidrule(r){3-5} \cmidrule(r){6-7}   
    LVIS~\cite{gupta2019lvis} \&    & part    & \cmark               & \cmark               &                           & 15,089  & 596,687  \\
    PACO~\cite{ramanathan2023paco}  & no\_part & \cmark               &   &                            & 103,178 & 2,062,536 \\
    \cmidrule(r){1-2} \cmidrule(r){3-5} \cmidrule(r){6-7}   
    \multirow{1}{*}{Entity-v2~\cite{qi2023high}}                   & cls     & \cmark               &   &                       & 31,913 & 579,076  \\
    \cmidrule(r){1-2} \cmidrule(r){3-5} \cmidrule(r){6-7}   
    \multirow{2}{*}{Pascal~\cite{everingham2010pascal}}     & train   & \cmark               & \cmark               &                          & 4,998   & 93,322   \\
                                & val   & \cmark               & \cmark               &                          & 5,105   & 95,462   \\
    \cmidrule(r){1-2} \cmidrule(r){3-5} \cmidrule(r){6-7}   
    MHP~\cite{li2017multiple}                         & train   &   & \cmark               &                           & 15,403  & 410,113  \\
    \cmidrule(r){1-2} \cmidrule(r){3-5} \cmidrule(r){6-7}   
    TextOCR~\cite{singh2021textocr}        & train      &   &   & \multicolumn{1}{c}{\cmark}             & 21,749          & 714,770          \\
    \cmidrule(r){1-2} \cmidrule(r){3-5} \cmidrule(r){6-7}   
    \rowcolor[HTML]{D8D8D8}
    \multicolumn{2}{c}{Total}   & \cmark               & \cmark               & \multicolumn{1}{c}{\cmark}           & 197,435         & 4,551,966        \\
    \bottomrule 
    \end{tabular}}
    \vspace{-10pt}
    \caption{Summary of the training datasets for Mask2Former+. Entity includes both thing and stuff categories. \vspace{-5pt}}    
    \label{tab:mask2former}
\end{table}

\paragraph{Model.}
Building on the foundation of the original Mask2Former~\cite{cheng2022masked}, we developed Mask2Former+, a panoptic segmentation model designed for multi-grained segmentation. 
We initialize our model from the Mask2Former checkpoint with the Swin-L backbone pre-trained on the COCO panoptic segmentation dataset~\cite{kirillov2019panoptic}. 
Besides the 200 entity queries that are trained for thing and stuff proposals, we added 50 additional expert queries for the segmenting parts and the visual text regions, respectively. 
Given that not all images have annotations for every type of segmentation (for instance, the TextOCR dataset provides annotations only for visual text regions), our model computes the group-wise matching loss exclusively for the annotations available in each dataset. 
This approach ensures that the model benefits from partial annotations without compromising its ability to recognize other levels of granularity when certain annotations are unavailable. 
Although most samples in our dataset also have semantic annotations such as object categories, we do not use them but only train the model for class-agnostic mask proposals. 
We train the model for 20k iterations on our combined segmentation dataset with a batch size of 16 using the Detectron2 library~\cite{wu2019detectron2}.\footnote{\url{https://github.com/facebookresearch/detectron2}}
 
\begin{table*}[!ht]
\centering
\vspace{-8pt}
    \scalebox{0.71}{
    \begin{tabular}{clcccccccccrc}
    \toprule
    \rowcolor[HTML]{C9DAF8}
    \multicolumn{3}{c}{Metadata}                                        & \multicolumn{3}{c}{Grounding Annotations}                             & \multicolumn{5}{c}{Semantic Granularity}                                                                              & \multicolumn{2}{c}{Data Size}                                \\
    \cmidrule(r){1-3} \cmidrule(r){4-6} \cmidrule(r){7-11} \cmidrule(r){12-13}
    \rowcolor[HTML]{D8D8D8}
    Task Type                & Dataset Name      & Image Source         & Mask                  & Box                   & Pointer               & Thing                 & Stuff                 & Part                  & Multi.                 & Text                  & \multicolumn{1}{c}{Train} & Val / Test                    \\
     \cmidrule(r){1-3} \cmidrule(r){4-6} \cmidrule(r){7-11} \cmidrule(r){12-13}
    \multirow{2}{*}{\shortstack[c]{Grd. Captioning\\(GCAP)}} & PNG               & COCO                 & \cmark & \cmark &                       & \cmark & \cmark &                       & \cmark &                       & 132,045                   & \multicolumn{1}{c}{8,435}   \\
                             & Flickr30K-Entity  & Flickr30K            &                       & \cmark &                       & \cmark & \cmark & \cmark & \cmark &                       & 148,915                   & 1,000 / 1,000                 \\
     \cmidrule(r){1-3} \cmidrule(r){4-6} \cmidrule(r){7-11} \cmidrule(r){12-13}
    \multirow{10}{*}{\shortstack[c]{Referential\\Expression\\Segmentation\\(RES)}}    & RefCOCO           & COCO                 & \cmark & \cmark &                       & \cmark &                       &                       &                       &                       & 113,311                   & -                           \\
                             & RefCOCO+          & COCO                 & \cmark & \cmark &                       & \cmark &                       &                       &                       &                       & 112,441                   & -                           \\
                             & RefCOCOg          & COCO                 & \cmark & \cmark &                       & \cmark &                       &                       &                       &                       & 80,322                    & -                           \\
                             & RefCLEF           & ImageCLEF            & \cmark & \cmark &                       & \cmark &                       &                       &                       &                       & 104,531                   & -                           \\
                             & gRefCOCO          & COCO                 & \cmark & \cmark &                       & \cmark &                       &                       &                       &                       & 194,233                   & -                           \\
                             & PhraseCut         & VG                   & \cmark & \cmark &                       & \cmark & \cmark & \cmark & \cmark &                       & 84,688                    & -                           \\
                             & D-Cube             & GRD                  & \cmark & \cmark &                       & \cmark &                       &                       & \cmark &                       & 9,499                     & -                           \\
                             & ReasonSeg         & OpenImages \& ScanNetV2 & \cmark & \cmark &                       & \cmark & \cmark & \cmark & \cmark &                       & 1,315                     & \multicolumn{1}{c}{344}     \\
                             & RIO               & COCO                 & \cmark & \cmark &                       & \cmark &                       &                       & \cmark &                       & 27,696                    & \multicolumn{1}{c}{34,170}  \\
                             & SK-VG             & VCR                  &                       & \cmark &                       & \cmark &                       &                       &                       &                       & 23,404                    & -                           \\
     \cmidrule(r){1-3} \cmidrule(r){4-6} \cmidrule(r){7-11} \cmidrule(r){12-13}
    \multirow{8}{*}{\shortstack[c]{Grounded\\Visual\\Question\\Answering\\(GVQA)}}     & VizWiz-Grounding  & VizWiz               & \cmark & \cmark &                       & \cmark & \cmark &                       &                       & \cmark & 6,494                     & 1,131 / 2,373                \\
                             & TextVQA-X         & OpenImages           & \cmark & \cmark &                       &                       &                       &                       &                       & \cmark & 14,476                    & \multicolumn{1}{c}{3,620}   \\
                             & GQA               & VG                   &                       & \cmark &                       & \cmark & \cmark & \cmark & \cmark &                       & 301,623                   & -                           \\
                             & VQS               & COCO                 &                       & \cmark &                       & \cmark &                       &                       &                       &                       & 20,380                    & \multicolumn{1}{c}{8,203}   \\
                             & Shikra-BinaryQA   & Flickr30K            &                       & \cmark &                       & \cmark & \cmark & \cmark & \cmark &                       & 4,044                     & \multicolumn{1}{c}{1,159}   \\
                             & EntityCount       & Entity-v2             & \cmark & \cmark &                       & \cmark & \cmark & \cmark & \cmark &                       & 11,088                    & \multicolumn{1}{c}{453}     \\
                             & FoodSeg-QA        & Recipe1M             & \cmark & \cmark &                       & \cmark &                       &                       & \cmark &                       & 7,114                     & -                           \\
                             & LVIS-QA           & COCO                 & \cmark & \cmark &                       & \cmark & \cmark &                       & \cmark &                       & 94,860                    & \multicolumn{1}{c}{3,611}   \\
     \cmidrule(r){1-3} \cmidrule(r){4-6} \cmidrule(r){7-11} \cmidrule(r){12-13}
    \multirow{16}{*}{\shortstack[c]{Referential\\Dialog\\(RD)}}     & RefCOCO-REG       & COCO                 & \cmark & \cmark & \cmark & \cmark &                       &                       &                       &                       & 17,395                    & -                           \\
                             & RefCOCO+-REG      & COCO                 & \cmark & \cmark & \cmark & \cmark &                       &                       &                       &                       & 17,383                    & -                           \\
                             & RefCOCOg-REG      & COCO                 & \cmark & \cmark & \cmark & \cmark &                       &                       &                       &                       & 22,057                    & -                           \\
                             & gRefCOCO-REG      & COCO                 & \cmark & \cmark & \cmark & \cmark &                       &                       &                       &                       & 20,282                    & -                           \\
                             & VG-SpotCap        & VG                   &                       & \cmark & \cmark & \cmark & \cmark & \cmark & \cmark &                       & 247,381                   & \multicolumn{1}{c}{232,935} \\
                             & V7W               & COCO                 &                       & \cmark & \cmark & \cmark &                       &                       &                       &                       & 22,805                    & 10,193 / 57,265               \\
                             & PointQA-Local     & VG                   &                       &                       & \cmark & \cmark &                       &                       &                       &                       & 27,426                    & 4,855 / 4,880                 \\
                             & PointQA-Twice     & VG                   &                       &                       & \cmark & \cmark &                       &                       &                       &                       & 36,762                    & 14,668 / 5,710                \\
                             & VCR-Open          & VCR                  &                       & \cmark & \cmark & \cmark &                       &                       &                       &                       & 58,340                    & -                           \\
                             & VCR-Multichoice   & VCR                  &                       & \cmark & \cmark & \cmark &                       &                       &                       &                       & 97,648                    & 26,534 / 25,263               \\
                             & ShikraRD          & Flickr30K            &                       & \cmark & \cmark & \cmark & \cmark & \cmark & \cmark &                       & 1,878                     & -                           \\
                             & SVIT-RD           & VG                   &                       & \cmark & \cmark & \cmark & \cmark & \cmark & \cmark &                       & 32,571                    & -                           \\
                             & Guesswhat-Guesser & COCO                 & \cmark & \cmark & \cmark & \cmark &                       &                       &                       &                       & 92,136                    & \multicolumn{1}{c}{19,665}  \\
                             & Guesswhat-Oracle  & COCO                 & \cmark & \cmark & \cmark & \cmark &                       &                       &                       &                       & 101,256                   & \multicolumn{1}{c}{21,643}  \\
                             & VG-RefMatch       & VG                   &                       & \cmark & \cmark & \cmark & \cmark & \cmark & \cmark &                       & 247,381                   & -                           \\
                             & HierText          & OpenImages           & \cmark & \cmark & \cmark &                       &                       &                       &                       & \cmark & 6,058                     & \multicolumn{1}{c}{3,885}  \\
    \bottomrule
    \end{tabular}}
\vspace{-10pt}
\caption{The full list of datasets used in M3G2. \vspace{-15pt}}
\label{tab:dataset}
\end{table*}

\vspace{-5pt}
\section{The M3G2 Dataset}
\label{app:dataset}
\vspace{-5pt}

In this section, we introduce the M3G2 dataset with \underline{M}ulti-\underline{M}odal \underline{M}ulti-\underline{G}rained \underline{G}rounding. 
M3G2 is a comprehensive dataset consisting of 36 sub-problems, derived and augmented from 27 existing datasets with grounded vision-language annotations. 
The dataset is categorized into four main types: (1) Grounded Image Captioning (GIC), (2) Grounded Visual Question Answering (GVQA), (3) Referential Expression Segmentation (RES), and (4) Referential Dialog (RD). 
Details on the dataset sources, image origins, types of grounding annotations, semantic granularity, and data statistics are summarized in Table~\ref{tab:dataset}. 
All datasets are formatted into the conversation format between a human user and a model assistant, where the user provides task objectives as instructions, and model responses are generated automatically based on the annotations.

\begin{figure*}
\centering
\begin{subfigure}[t]{1.0\textwidth}
\centering
\includegraphics[width=1.0\linewidth]{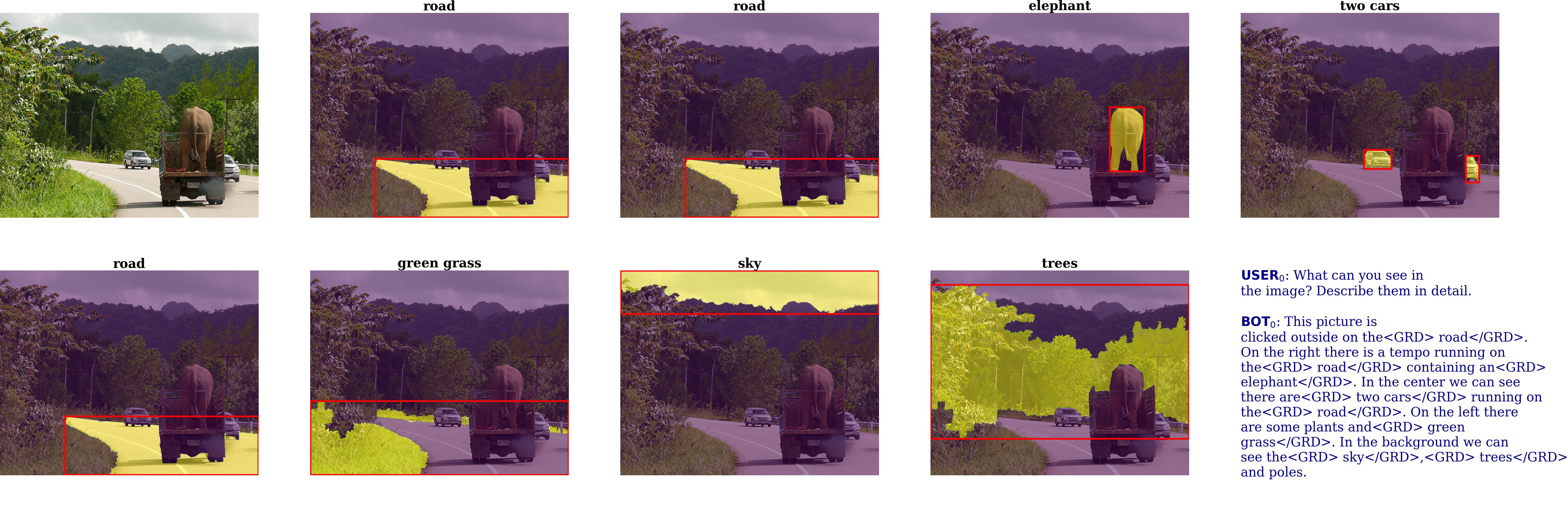}
\vspace{-10pt}
\caption{Example of the detailed captioning task originated from PNG.}
\label{fig:data_sample_png_1}
\end{subfigure}
~
\begin{subfigure}[t]{1.0\textwidth}
\centering
\includegraphics[width=1.0\linewidth]{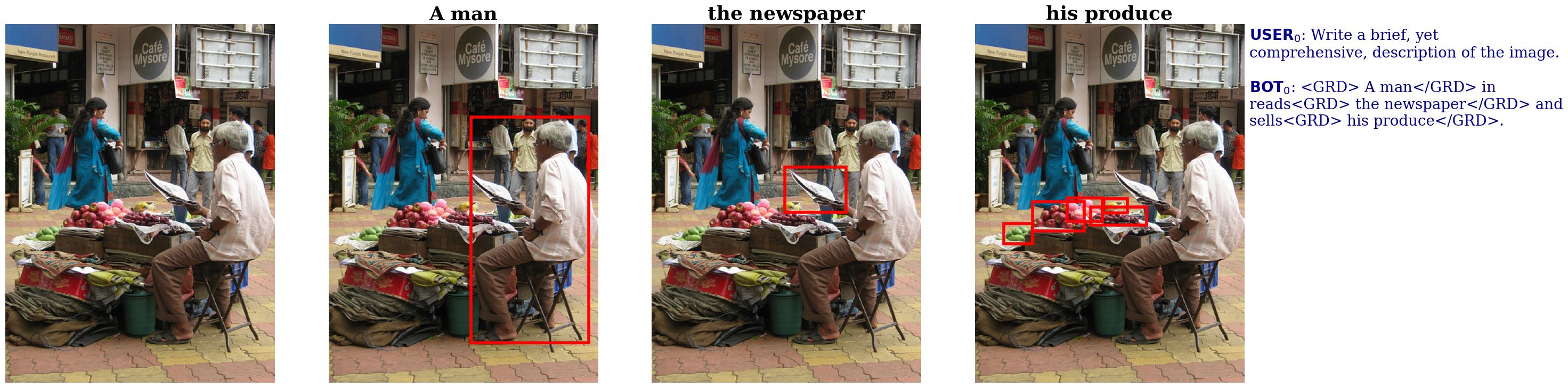}
\vspace{-10pt}
\caption{Example of the short captioning task originated from Flickr30K-Entity.}
\label{fig:data_sample_flickr30k_1}
\end{subfigure}
\caption{Data samples for the GCAP task.}
\label{fig:data_sample_captioning}
\end{figure*}

\paragraph{Grounded Image Captioning (GIC).} 
GIC focuses on generating image captions that ground to visual entities presented in the image. 
We incorporate the Panoptic Narrative Grounding (PNG)~\cite{kirillov2019panoptic} and Flickr30K-Entity~\cite{plummer2015flickr30k} datasets. 
PNG, derived from Localize Narrative~\cite{pont2020connecting} and COCO Segmentation~\cite{caesar2018coco}, provides long and detailed narratives with an average of 36.5 words per description, exemplified in Figure~\ref{fig:data_sample_png_1}. 
These narratives are rich in detail, offering a high coverage of the visual content including the background. Flickr30K-Entity, offering concise captions with box annotations, complements PNG with its larger vocabulary and finer granularity, as shown in \ref{fig:data_sample_flickr30k_1}. The example instruction templates used to construct the conversation are listed in Table~\ref{tab:prompt_caption}, where we use key words such as "short/briefly" and "in detail" to distinguish between short and long captioning.

\begin{table}[!t]
    \centering
    \resizebox{1.0\columnwidth}{!}{
    \begin{tabular}{ll}
    \toprule
    \rowcolor[HTML]{D8D8D8}
    Instruction Templates for Brief Captioning & Instruction Templates for Detailed Captioning \\
    \cmidrule(r){1-1}\cmidrule(r){2-2}
    Describe the image briefly. & Describe the image in detail.\\
    Describe the image in a few words. &  Describe the picture's every detail. \\
    Describe the image in a short sentence. & Describe the given picture in very detail. \\
    Describe the image in a clear and concise manner. & Make a fine description of the image. \\
    Generate a short caption for the picture. & Generate a long caption for the given image. \\
    Caption the image in a few words. & Give me a detailed caption of this image. \\
    \bottomrule 
    \end{tabular}}
    \vspace*{-10pt}
    \caption{Instruction templates for the GIC task. \vspace*{-15pt}}
    \label{tab:prompt_caption}
\end{table}

\begin{figure*}
\centering
\begin{subfigure}[t]{1.0\textwidth}
\centering
\includegraphics[width=1.0\linewidth]{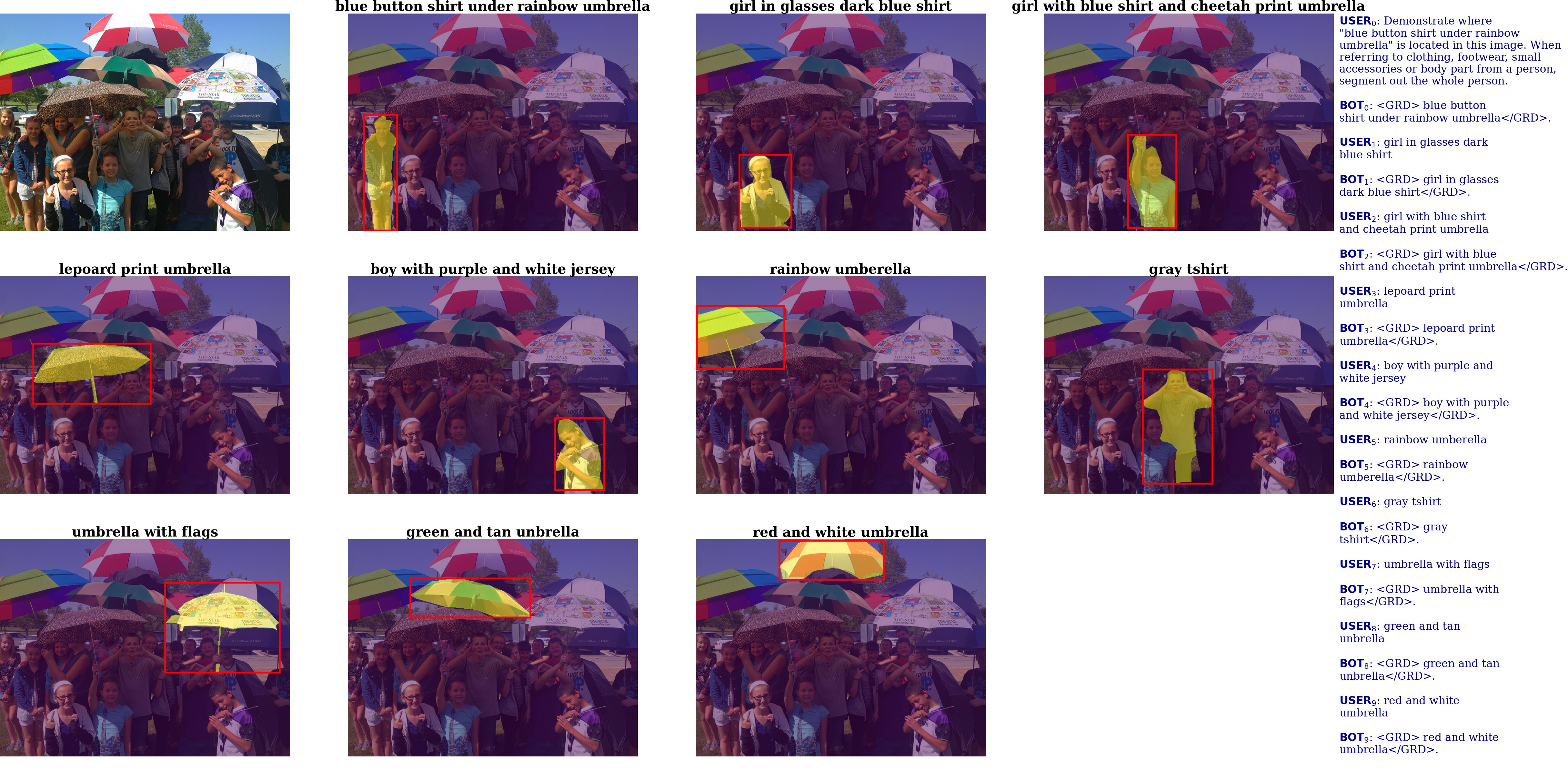}
\vspace{-10pt}
\caption{Example of the referential expression segmentation task originated from RefCOCO+.}
\label{fig:data_sample_refcoco+_1}
\end{subfigure}
~
\begin{subfigure}[t]{1.0\textwidth}
\centering
\includegraphics[width=1.0\linewidth]{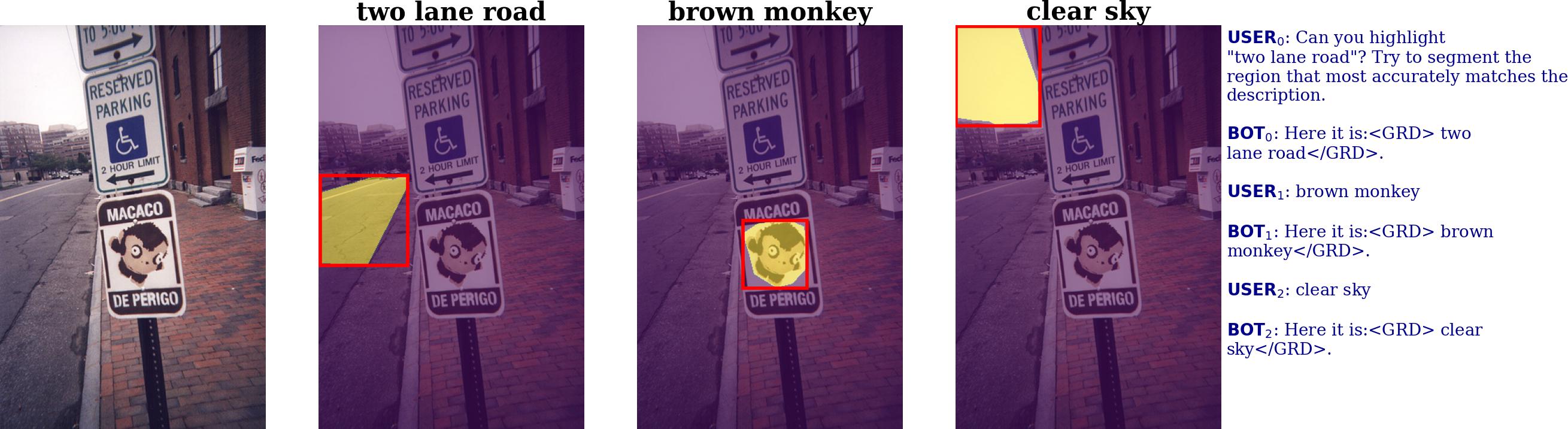}
\vspace{-10pt}
\caption{Example of the referential expression segmentation task originated from PhraseCut.}
\label{fig:data_sample_phrasecut_1}
\end{subfigure}
~
\begin{subfigure}[t]{1.0\textwidth}
\centering
\includegraphics[width=1.0\linewidth]{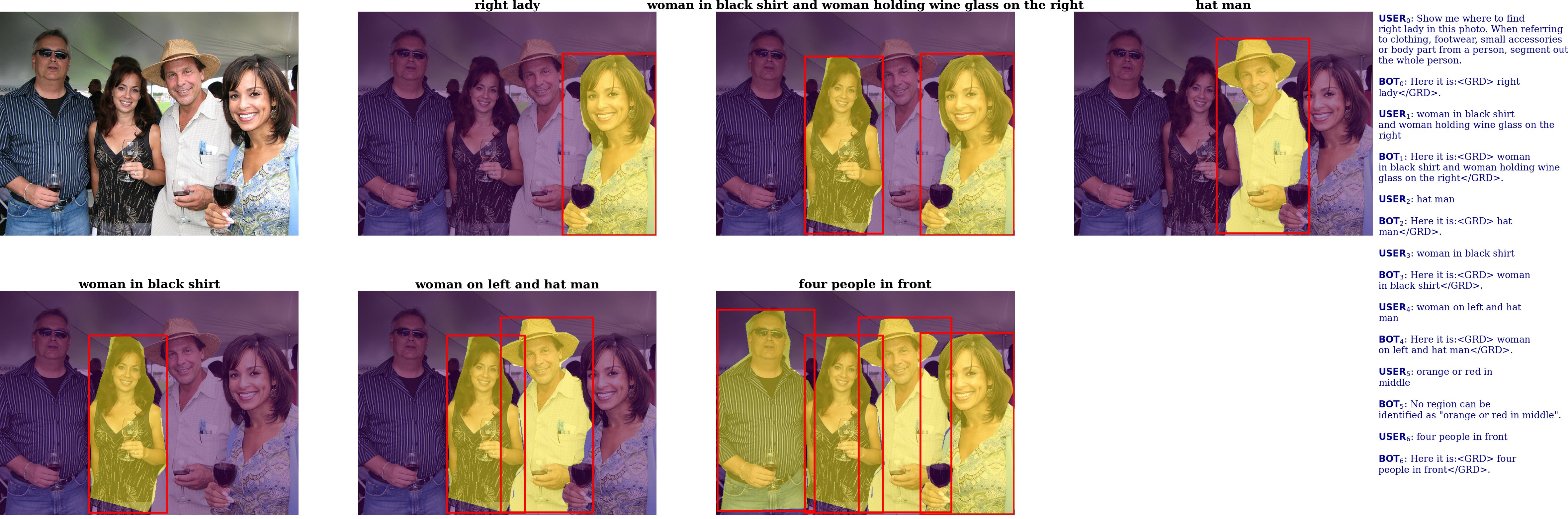}
\vspace{-10pt}
\caption{Example of the generalized referential expression segmentation task originated from gRefCOCO.}
\label{fig:data_sample_gres_1}
\end{subfigure}
\caption{Data samples for the RES task (part 1).}
\label{fig:data_sample_res1}
\end{figure*}

\begin{figure*}
\centering
\begin{subfigure}[t]{1.0\textwidth}
\centering
\includegraphics[width=1.0\linewidth]{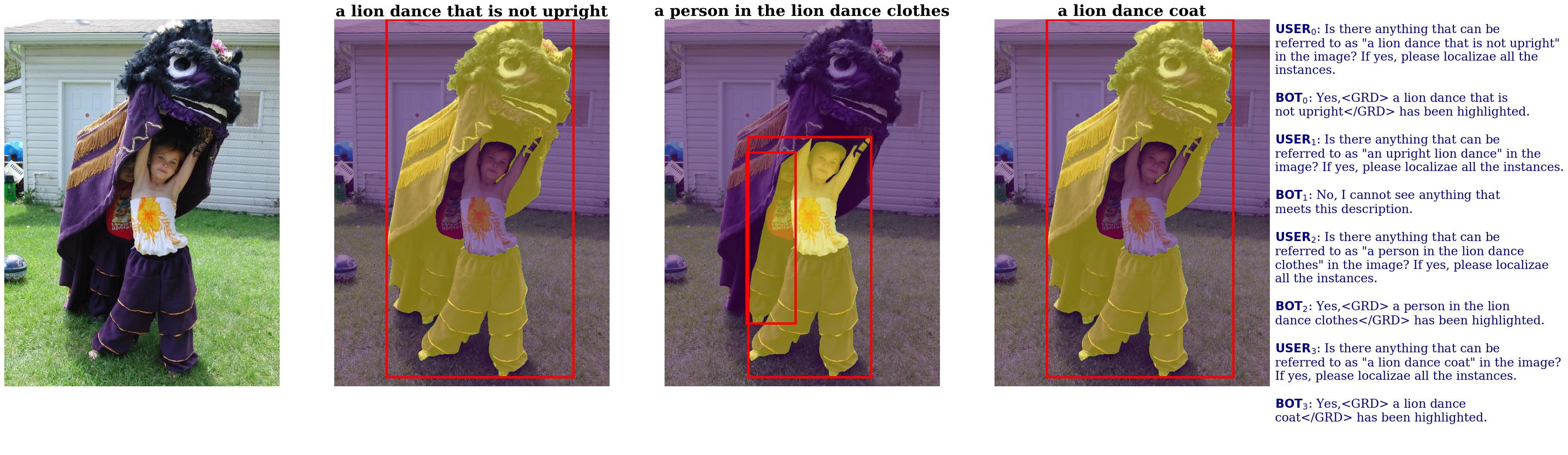}
\vspace{-10pt}
\caption{Example of the generalized referential expression segmentation task originated from D-Cube.}
\label{fig:data_sample_dcube_1}
\end{subfigure}
~
\begin{subfigure}[t]{1.0\textwidth}
\centering
\includegraphics[width=1.0\linewidth]{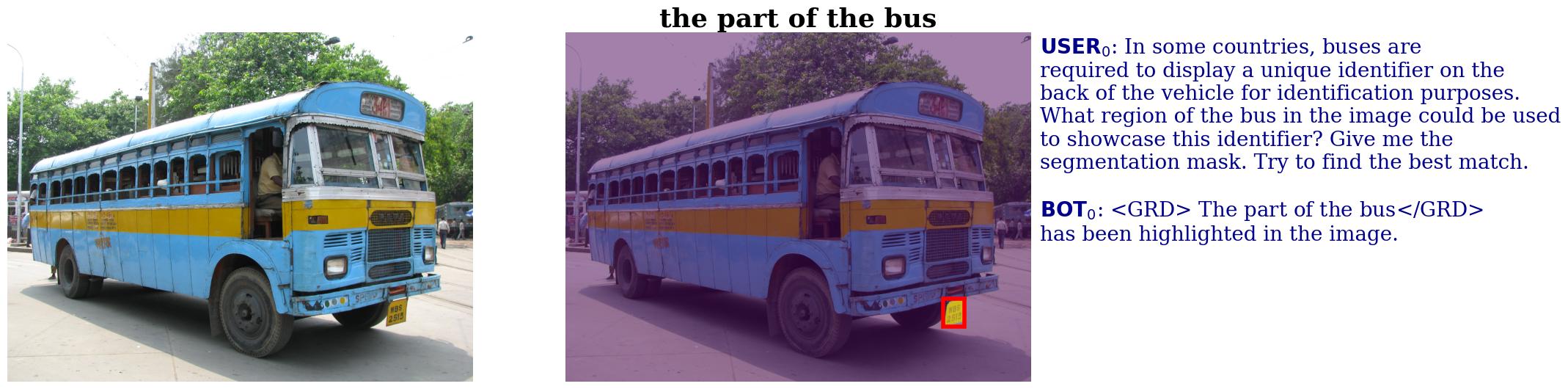}
\vspace{-10pt}
\caption{Example of the reasoning-based referential expression segmentation task originated from ReasonSeg.}
\label{fig:data_sample_reasonseg_1}
\end{subfigure}
~
\begin{subfigure}[t]{1.0\textwidth}
\centering
\includegraphics[width=1.0\linewidth]{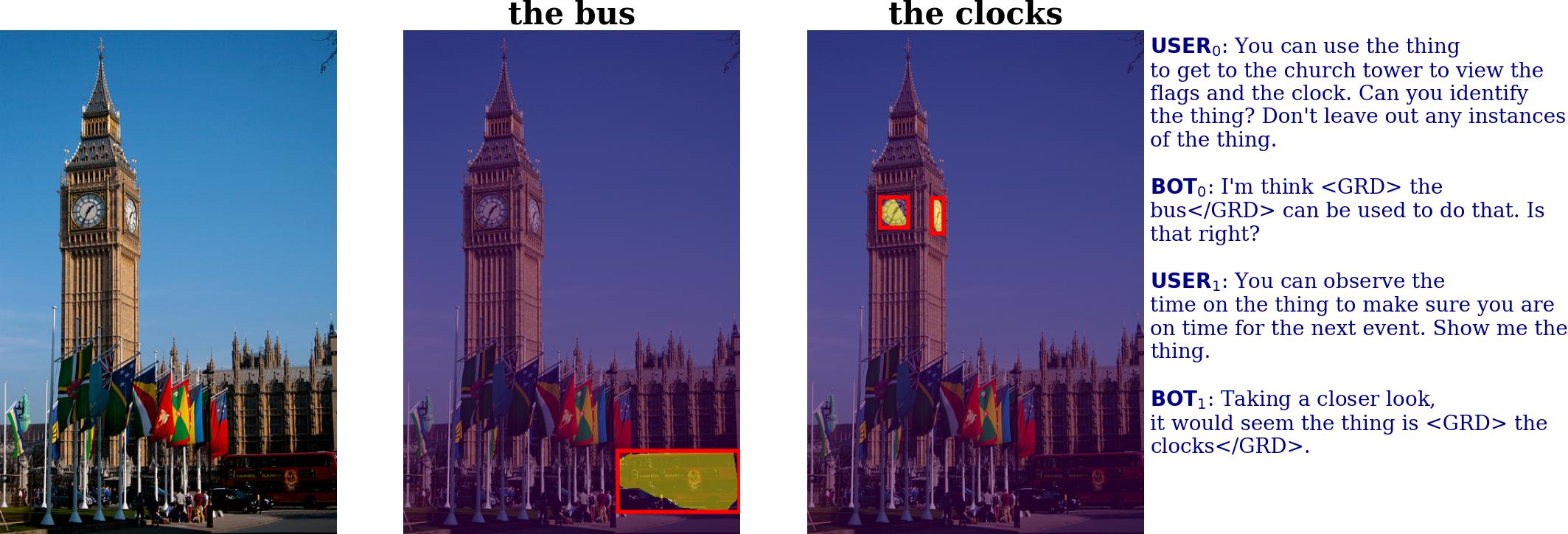}
\vspace{-10pt}
\caption{Example of the reasoning-based referential expression segmentation task originated from RIO.}
\label{fig:data_sample_rio_1}
\end{subfigure}
~
\begin{subfigure}[t]{1.0\textwidth}
\centering
\includegraphics[width=1.0\linewidth]{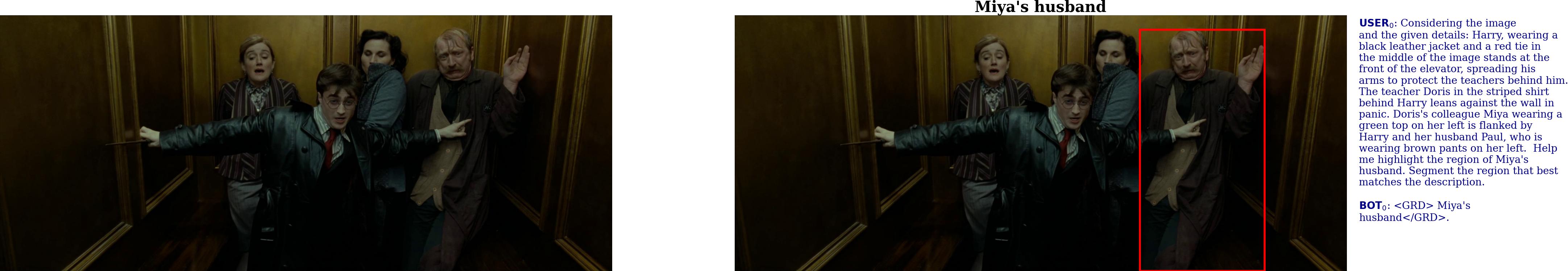}
\vspace{-10pt}
\caption{Example of the reasoning-based referential expression segmentation task originated from SK-VG.}
\label{fig:data_sample_skvg_1}
\end{subfigure}
\caption{Data samples for the RES task (part 2).}
\label{fig:data_sample_res2}
\end{figure*}

\vspace{-5pt}
\paragraph{Referential Expression Segmentation (RES).} 
RES is a task combining language understanding with precise visual segmentation. 
Our dataset includes 10 diverse sources. 
To improve the learning efficiency and enhance contextual understanding, we format queries from the same image into a simulated multi-turn dialog, as illustrated in Figures \ref{fig:data_sample_res1} and \ref{fig:data_sample_res2}. 
We employ the widely used RefCOCO/+/g datasets~\cite{kazemzadeh2014referitgame,mao2016refcocog} and RefCLEF~\cite{rohrbach2016refclef} for single-object RES. 
gRefCOCO~\cite{liu2023gres} is employed for multi-object and negative queries. 
To enhance the visual diversity, we also incorporate PhraseCut~\cite{wu2020phrasecut} and D-Cube~\cite{xie2023described} that use an image source different than COCO. 
Additionally, ReasonSeg~\cite{lai2023lisa}, RIO~\cite{qu2023rio}, and SK-VG \cite{wu2023advancing} are included, where a textual context is given and the models need to not only understand that context, but also equips with a certain degree of commonsense knowledge to successfully solve the query, such as shown in Figure~\ref{fig:data_sample_reasonseg_1}, \ref{fig:data_sample_rio_1} and \ref{fig:data_sample_skvg_1}. 
The dialogue templates are listed in Table~\ref{tab:prompt_res}.

\begin{table}[!t]
    \centering
    \resizebox{1.0\columnwidth}{!}{
    \begin{tabular}{ll}
    \toprule
    \rowcolor[HTML]{D8D8D8}
    \multicolumn{2}{c}{Instruction Templates For RES} \\
    \cmidrule(r){1-2}
    Highlight ``\{\}'' in the image. &
    Segment ``\{\}'' in the image. \\
    Segment: \{\}.  &
    Help me segment out \{\}. \\
    Localize ``\{\}'' in the image.  &
    Help me localize \{\}. \\
    Help me highlight the region of \{\}.  &
    Demonstrate where ``\{\}'' is located in this image. \\
    Show me where to find {} in this photo.  &
    Identify and mark the region of \{\} for me. \\
    Can you highlight ``\{\}''?  &
    Can you extract the segment: \{\} for me? \\
    Can you localize ``\{\}'' in this image?  &
    Could you please segment out \{\} in the image? \\
    \bottomrule 
    \end{tabular}}
    \vspace*{-10pt}
    \caption{Templates used for the RES task. \vspace*{-15pt}}
    \label{tab:prompt_res}
\end{table}

\begin{figure*}
\centering
\begin{subfigure}[t]{.48\textwidth}
\centering
\includegraphics[width=1.0\linewidth]{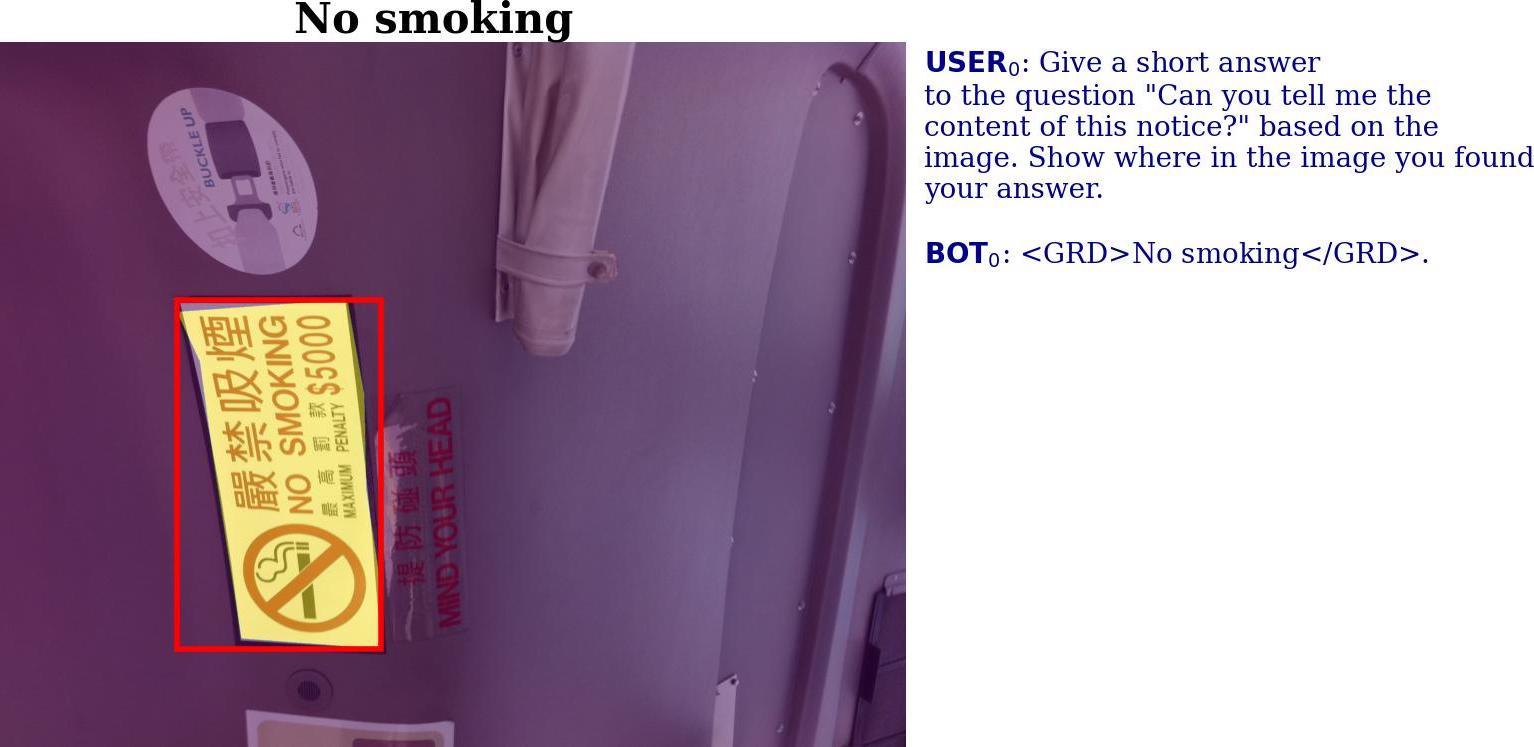}
\vspace{-10pt}
\caption{Example of grounded VQA originated from VizWiz-Grounding.}
\label{fig:data_sample_vizwiz_1}
\end{subfigure}
~
\begin{subfigure}[t]{.48\textwidth}
\centering
\includegraphics[width=1.0\linewidth]{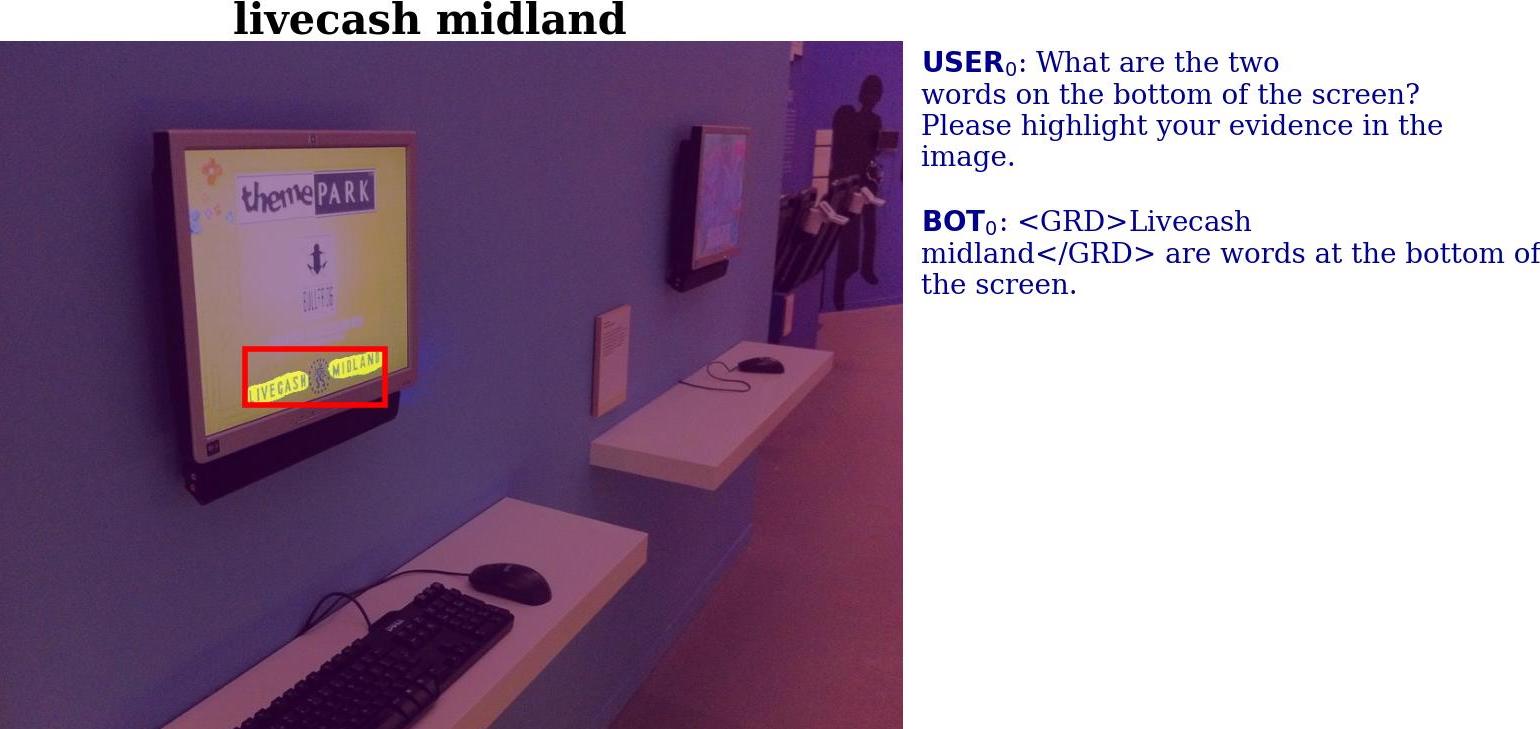}
\vspace{-10pt}
\caption{Example of grounded VQA originated from TextVQA-X.}
\label{fig:data_sample_textvqax_1}
\end{subfigure}
~
\begin{subfigure}[t]{1.0\textwidth}
\centering
\includegraphics[width=1.0\linewidth]{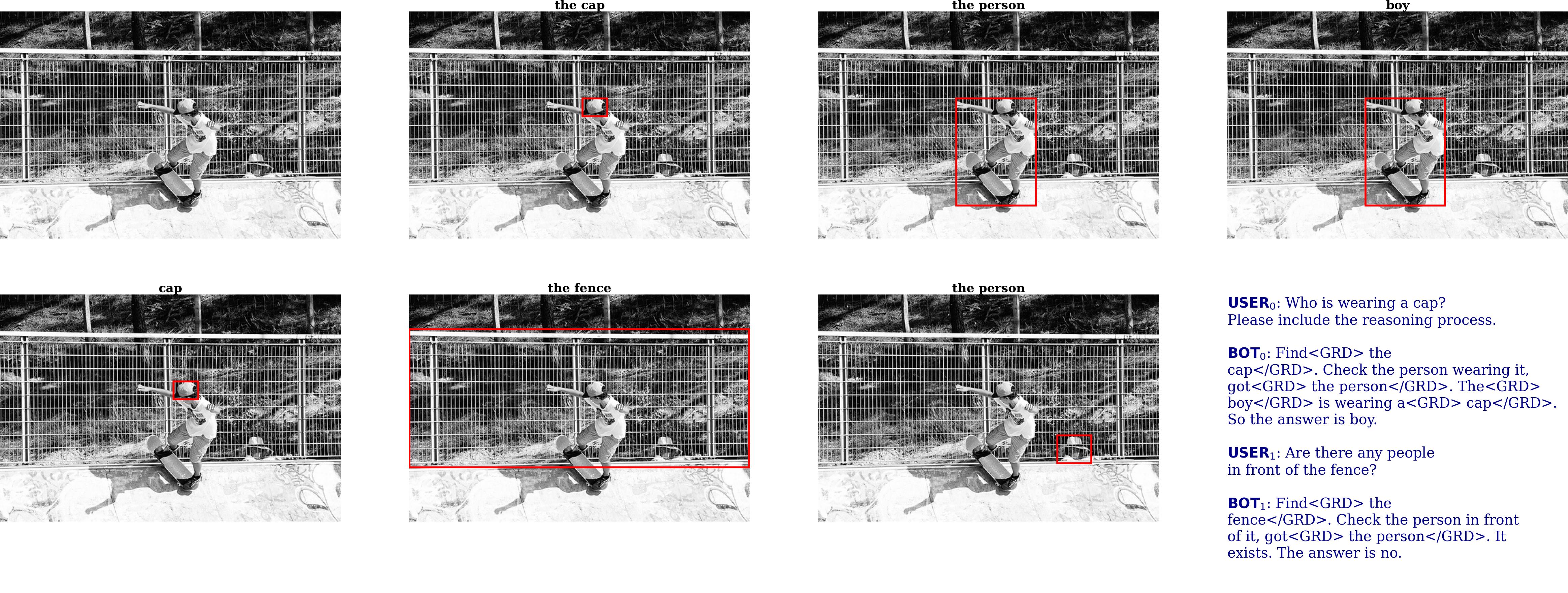}
\vspace{-10pt}
\caption{Example of grounded VQA with chain-of-thought reasoning originated from GQA.}
\label{fig:data_sample_gqa_1}
\end{subfigure}
~
\begin{subfigure}[t]{1.0\textwidth}
\centering
\includegraphics[width=1.0\linewidth]{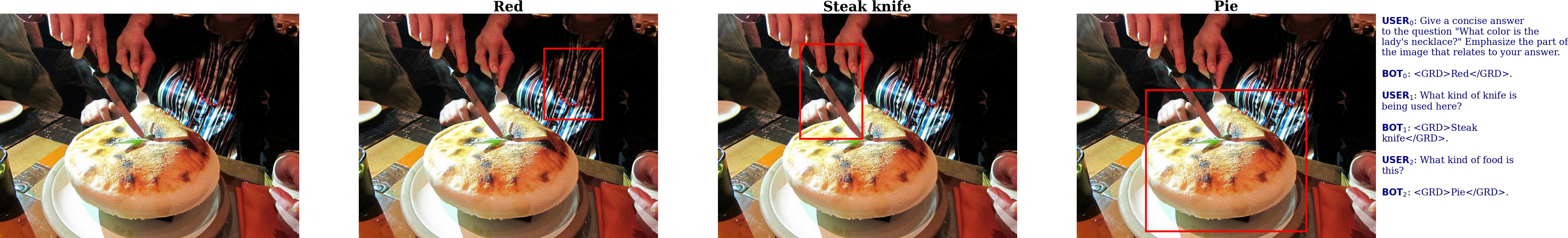}
\vspace{-10pt}
\caption{Example of grounded VQA task originated from VQS.}
\label{fig:data_sample_vqs_1}
\end{subfigure}
~
\begin{subfigure}[t]{1.0\textwidth}
\centering
\includegraphics[width=1.0\linewidth]{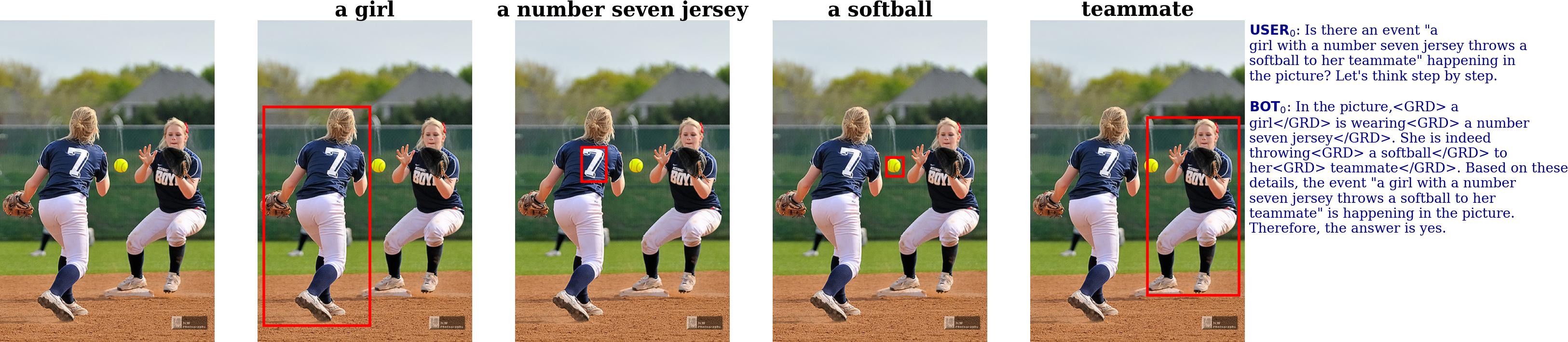}
\vspace{-10pt}
\caption{Example of the grounded VQA with chain-of-thought reasoning originated from Shikra-BinaryQA.}
\label{fig:data_sample_shikra_1}
\end{subfigure}
\caption{Data samples of the GVQA task (part 1).}
\label{fig:data_sample_vqa1}
\end{figure*}

\begin{figure*}
\centering
\begin{subfigure}[t]{0.97\textwidth}
\centering
\includegraphics[width=1.0\linewidth]{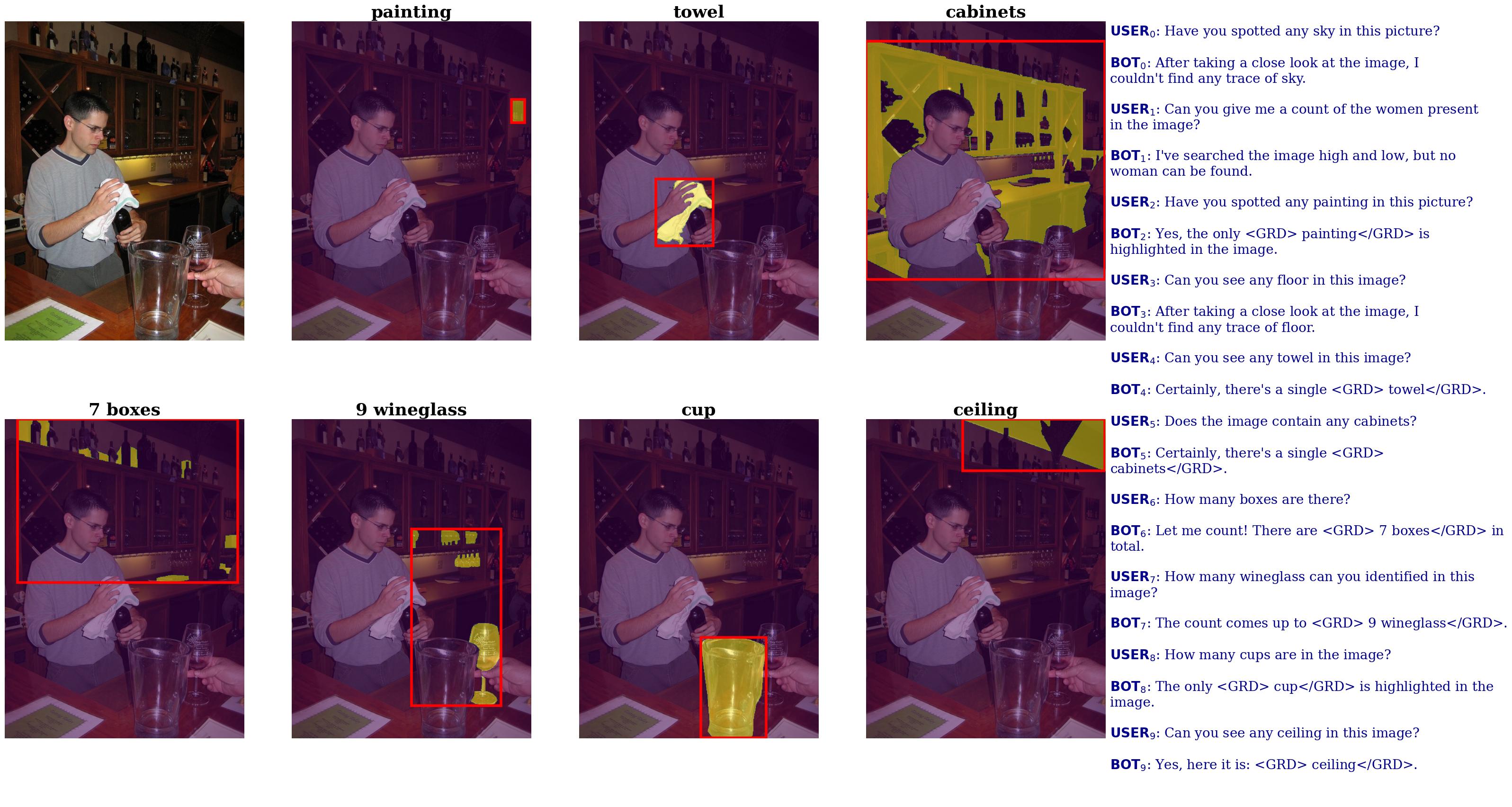}
\vspace{-30pt}
\caption{Example of the grounded VQA task originated from EntityCount. }
\label{fig:data_sample_entitycnt_1}
\end{subfigure}
~
\begin{subfigure}[t]{1.0\textwidth}
\centering
\includegraphics[width=1.0\linewidth]{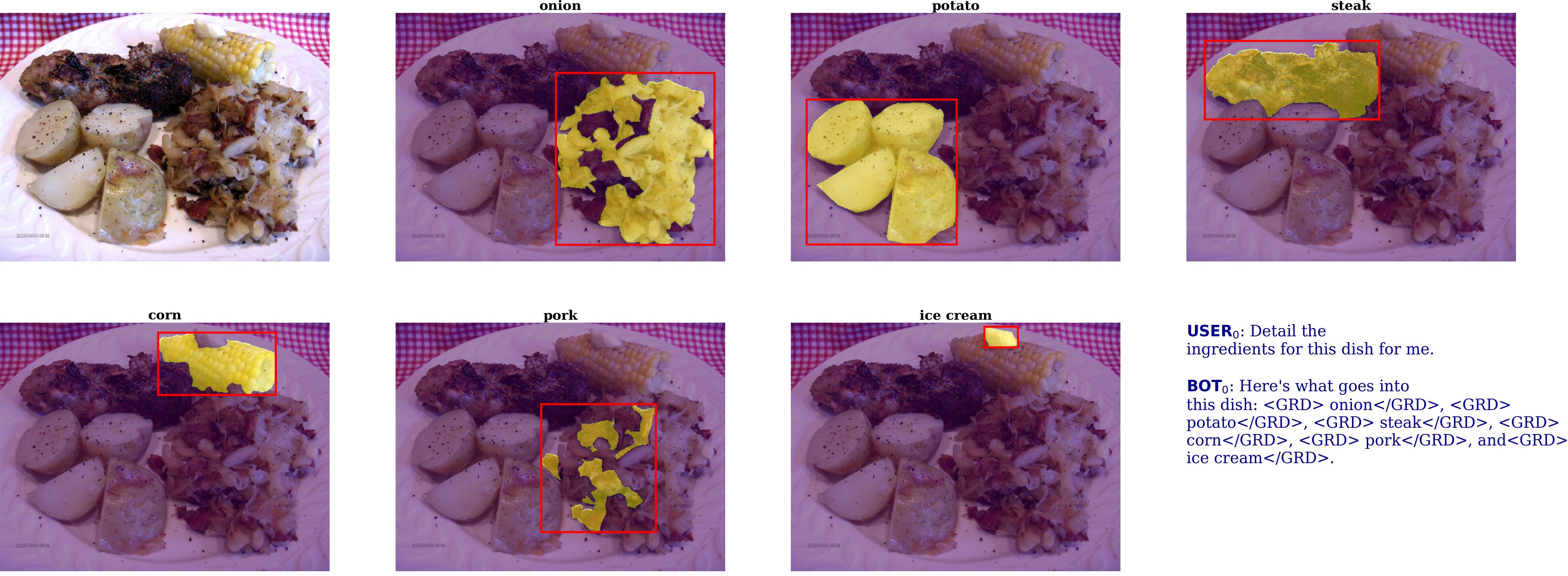}
\vspace{-10pt}
\caption{Example of the grounded VQA task originated from FoodSeg-QA.}
\label{fig:data_sample_foodseg_1}
\end{subfigure}
~
\begin{subfigure}[t]{1.0\textwidth}
\centering
\includegraphics[width=1.0\linewidth]{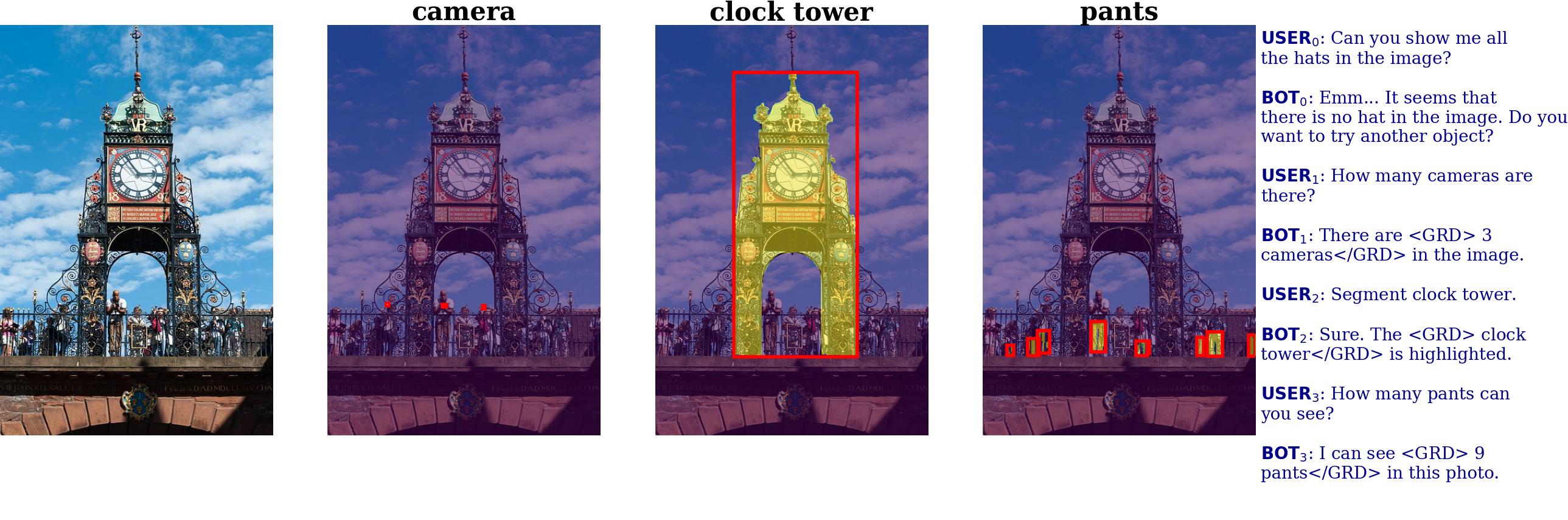}
\vspace{-35pt}
\caption{Example of the grounded VQA task originated from LVIS-QA.}
\label{fig:data_sample_lvis_1}
\end{subfigure}
\vspace{-15pt}
\caption{Data samples of the GVQA task (part 2).}
\label{fig:data_sample_vqa2}
\end{figure*}

\vspace{-5pt}
\paragraph{Grounded Visual Question Answering (GVQA).} 
The GVQA task extends the visual question answering by additionally requiring visual grounding of the answer.
We include 8 datasets for the grounded VQA task in M3G2. 
First, we collect and organize some existing datasets that can directly fit into our grounded vision-language task framework, including VizWiz-Grounding~\cite{chen2022vizwizg}, TextVQA-X~\cite{rao2021first}, GQA~\cite{hudson2019gqa}, VQA~\cite{gan2017vqs} and Shikra-BinaryQA~\cite{chen2023shikra} (Figure~\ref{fig:data_sample_vqa1}). 
To further improve the data scale and visual concept coverage, we enlarge the GVQA collection by re-purposing existing panoptic segmentation datasets with templated instructions and model responses. 
Specifically, based on the annotations from LVIS~\cite{gupta2019lvis} and EntityV2~\cite{qi2023high}, we design questions about object presence, object counting, and segment query with a possibly negative request (i.e. the target object does not exist in the image), for the model to learn to recognize a diverse set of concepts more faithfully. 
See Figure~\ref{fig:data_sample_vqa2} for examples of such multi-turn QA, and example question templates used in Table~\ref{tab:prompt_vqa}. 

\begin{table}[!t]
    \centering
    \resizebox{0.95\columnwidth}{!}{
    \begin{tabular}{l}
    \toprule
    \rowcolor[HTML]{D8D8D8}
    Instruction Templates For Short Response VQA. \\
    \{\}Answer with a single word or a short phrase. \\
    Given the image, answer the question ''\{\}'' with a single word or a short phrase. \\
    Give a short answer to the question ''\{\}'' based on the image. \\
    \midrule
    \rowcolor[HTML]{D8D8D8}
    Instruction Templates For Chain-of-Thought Response VQA. \\
    \{\}Let's think step by step. \\
    \{\}Please include the reasoning process. \\
    \{\}Before giving the answer, please explain your reasoning. \\
    \{\}Explain your logic before giving the answer. \\
    Please answer the following question ''\{\}'', and describe your thought process. \\
    \midrule
    \rowcolor[HTML]{D8D8D8}
    Instruction Templates For Grounding Answer to Masks. \\
    Show where in the image you found your answer. \\
    Mark the part of the image that supports your answer. \\
    Please highlight your evidence in the image. \\
    Point out the evidence from the image. \\
    Indicate the area in the image that justifies your response. \\
    Highlight the section of the image that backs up your answer. \\
    Shade the section of the image that confirms your reply. \\
    Emphasize the part of the image that relates to your answer. \\
    \midrule
    \rowcolor[HTML]{D8D8D8}
    Instruction Templates For Object Presence QA. \\
    Is \{\} present in the image? \\
    Is there any \{\} in this image? \\
    \rowcolor[HTML]{D8D8D8}
    Instruction Templates For Object Counting QA. \\
    How many \{\} can you see in this image? \\
    Count the number of \{\}. \\
    \midrule
    \rowcolor[HTML]{D8D8D8}
    Instruction Templates For Object Segmentation Request. \\
    Segment \{\}. \\
    Highlight all the \{\} in this image. \\
    Show me all the \{\} presented in the picture. \\
    \bottomrule 
    \end{tabular}}
    \vspace*{-7pt}
    \caption{Templates used for the GVQA task. \vspace*{-16pt}}
    \label{tab:prompt_vqa}
\end{table}

\begin{figure*}
\centering
\begin{subfigure}[t]{1.0\textwidth}
\centering
\includegraphics[width=1.0\linewidth]{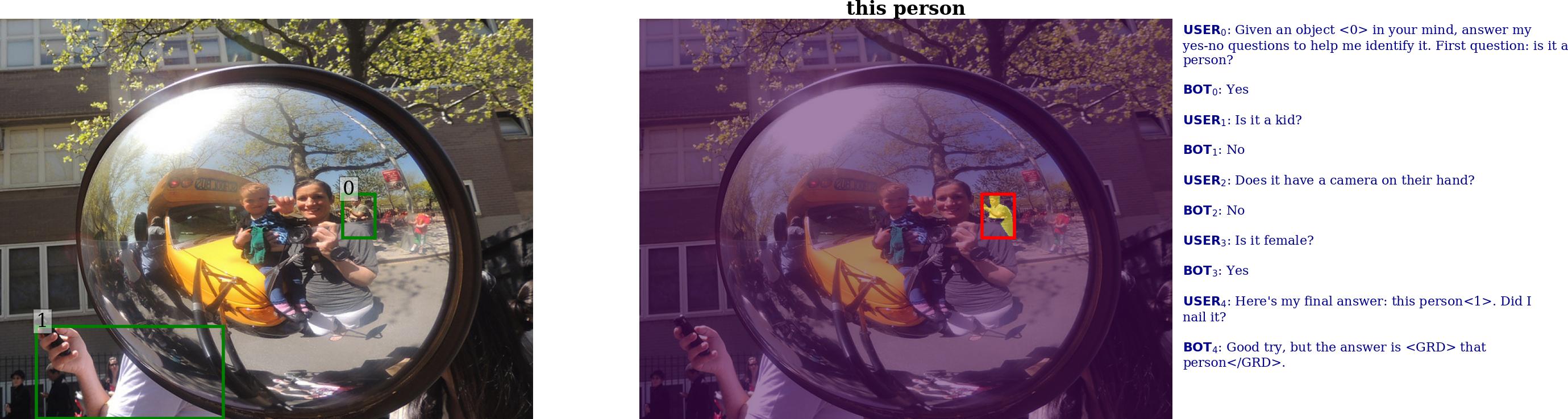}
\vspace{-10pt}
\caption{Example of the guesswhat game originated from GuessWhat.}
\label{fig:data_sample_guesswhat_1}
\end{subfigure}
~
\begin{subfigure}[t]{1.0\textwidth}
\centering
\includegraphics[width=1.0\linewidth]{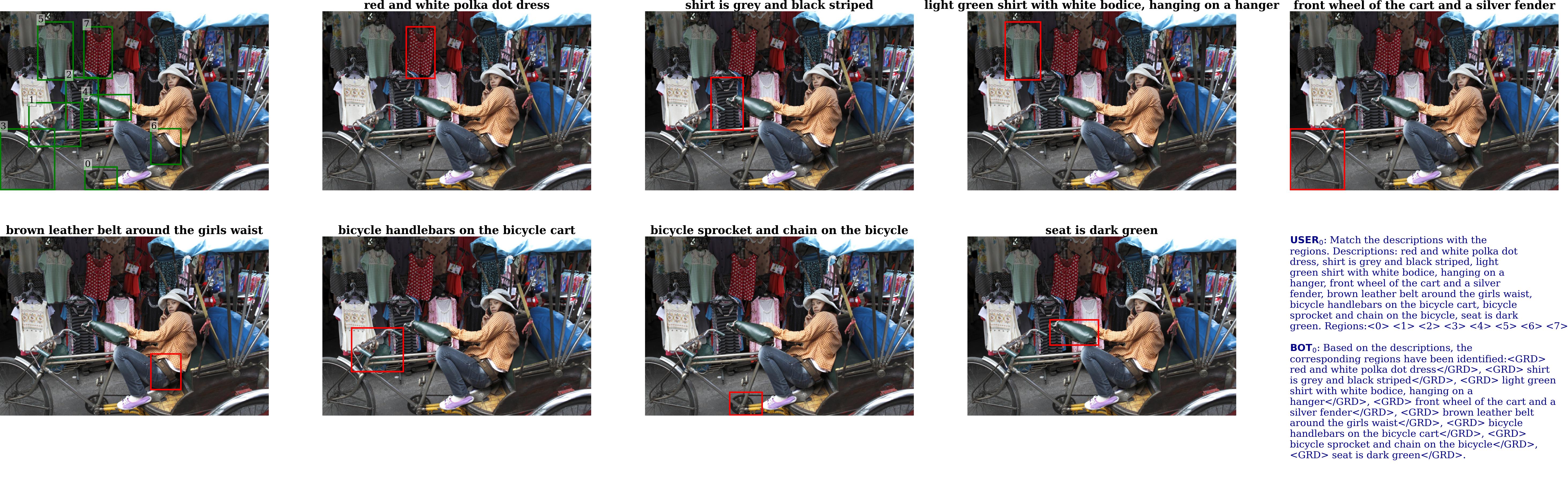}
\vspace{-10pt}
\caption{Example of the referential region matching game originated from VG.}
\label{fig:data_sample_vg_2}
\end{subfigure}
~
\begin{subfigure}[t]{1.0\textwidth}
\centering
\includegraphics[width=1.0\linewidth]{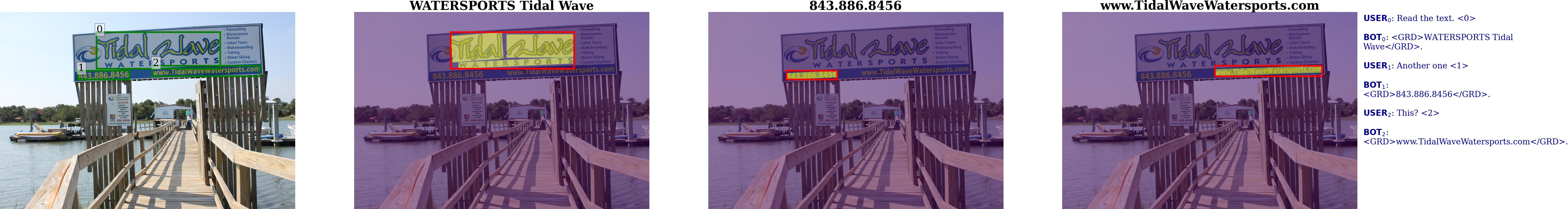}
\vspace{-10pt}
\caption{Example of the referred text reading task originated from HierText.}
\label{fig:data_sample_hiertext_1}
\end{subfigure}
\caption{Data samples of the RD task (part 2).}
\label{fig:data_sample_rd2}
\end{figure*}

\vspace{-5pt}
\paragraph{Referential Dialog (RD)}. 
RD features multi-modal conversations where the user can refer to objects or regions in the image by a spatial prompt (e.g. a bounding box).
We include various types of RD in our dataset and the templates used are listed in Table~\ref{tab:prompt_rd}. 
First, we add several existing RD datasets such as V7W~\cite{zhu2016visual7w}, PointQA~\cite{mani2020point}, VCR~\cite{zellers2019recognition}, ShikraRD~\cite{chen2023shikra} and SVIT~\cite{zhao2023svit} without much modifications.
We then revisit the RefCOCO series~\cite{yu2016modeling,mao2016generation,liu2023gres} for referential expression generation, where the referred object is given and the goal is to generate a unique description that leads to that object. 
We use the region caption annotations from the VG dataset~\cite{krishna2017visual} for region captioning and a region-matching game.
We select a set of region pointers and several descriptions to provide to the model, and the goal is to match the pointed regions with the descriptions (Figure~\ref{fig:data_sample_vg_2}).
We repurpose the GuessWhat dataset~\cite{de2017guesswhat} to make it fit into our RD formulation, as shown in Figure~\ref{fig:data_sample_guesswhat_1}. 
We also construct a referred text reading task based on the HierText~\cite{long2022towards} dataset and enhance the model's capability of text recognition, as shown in Figure~\ref{fig:data_sample_hiertext_1}. 

\begin{table}[!t]
    \centering
    \resizebox{0.95\columnwidth}{!}{
    \begin{tabular}{l}
    \toprule
    \rowcolor[HTML]{D8D8D8}
    Instruction Templates For REG. \\
    Provide a distinct description for that $<$PTR$>$ \\
    Describe the selected area in a unique way. $<$PTR$>$ \\
    Share a unique description of the region $<$PTR$>$ \\
    Offer a one-of-a-kind descriptor $<$PTR$>$ \\
    Describe the selected area $<$PTR$>$ uniquely. \\
    Point out $<$PTR$>$ in the picture with a unique description. \\
    Tell me how $<$PTR$>$ stands out in the photo. \\
    Use your words to highlight just $<$PTR$>$ in the image. \\
    Please describe $<$PTR$>$ in the image in a way that it can be uniquely identified. \\
    If you had to describe just $<$PTR$>$ to someone, how would you do it? \\
    What makes $<$PTR$>$ different from everything else in the picture? \\
    How can you describe $<$PTR$>$ in the image in a way that it can be uniquely identified? \\
    Can you provide a referring expression for $<$PTR$>$ such that it sets it apart from others? \\
    Let's play a game! Describe $<$PTR$>$ in the photo so I can find it. \\
    \midrule
    \rowcolor[HTML]{D8D8D8}
    Instruction Templates For Region Captioning. \\
    Describe it $<$PTR$>$. \\
    Describe the region $<$PTR$>$ in a few words. \\
    Describe the region $<$PTR$>$ in a short phrase. \\
    Describe the selected area $<$PTR$>$. \\
    Provide a brief description of this part $<$PTR$>$. \\
    Give a short caption for this $<$PTR$>$. \\
    Provide a brief description of the area marked $<$PTR$>$. \\
    Tell me about the contents in the selected zone $<$PTR$>$. \\
    Provide a concise description for this spot $<$PTR$>$. \\
    Narrate what you see in the indicated area $<$PTR$>$. \\
    What is in the region $<$PTR$>$? Describe in a phrase. \\
    What can you see in this area $<$PTR$>$? \\
    How would you describe the content at $<$PTR$>$? \\
    How would you caption this particular region $<$PTR$>$? \\
    What's depicted in the marked area $<$PTR$>$? \\
    \bottomrule 
    \end{tabular}}
    \vspace*{-7pt}
    \caption{Templates used for the RD task. \vspace*{-10pt}}
    \label{tab:prompt_rd}
\end{table}

\vspace{-5pt}
\section{\model~Implementation Details}
\label{app:implement}


\paragraph{Data Balancing.}

In constructing the M3G2 dataset, we recognized the need to address the varying scales of the multiple constituent datasets to ensure a balanced data distribution during training.
To achieve this, we have implemented dataset-specific sampling strategies, adjusting the volume of data from each source dataset through either up-sampling or down-sampling. The ratios we applied are as follows:
\begin{itemize}
    \item PNG: up-sampled by a factor of 2.
    \item Flickr30k-Entities: up-sampled by 1.5 times.
    \item RefCOCO$^{+}$: up-sampled by 1.5 times.
    \item RefCOCOg: up-sampled by 1.5 times.
    \item SK-VG: up-sampled by a factor of 2.
    \item Dcube (multiturn): up-sampled by a factor of 10. 
    \item ReasonSeg: up-sampled by a factor of 10.
    \item Shikra-Binary: up-sampled by a factor of 10. 
    \item VCR-Open (multiturn): down-sampled by half.
    \item VCR-Multiturn: down-sampled to 10\%.
    \item VizWiz: up-sampled by a factor of 3.
    \item LVIS-QA: down-sampled by half.
    \item TextVQAX: up-sampled by a factor of 2.
    \item EntityCount: up-sampled by a factor of 2.
    \item VG-SpotCap: down-sampled by half.
    \item Shikra-RD: up-sampled by a factor of 10.
    \item HierText: up-sampled by a factor of 5.
    \item GuessWhat-Oracle: down-sampled to 20\%.
    \item GuessWhat-Guesser: down-sampled to 20\%.
    \item SVIT: up-sampled by a factor of 3.
\end{itemize}
\vspace{5pt}
The balanced sampled dataset contains 1.8 million samples in total.

\paragraph{Learning from Both Box and Mask Supervision.} 
In the M3G2 dataset, not all sub-datasets include mask supervision, necessitating a hybrid loss approach to effectively benefit from grounded supervision from both mask and box annotations. 
We address this by employing different loss functions based on the type of annotation available. 
When mask annotations are available, we apply the dice loss $\mathcal{L}_{\text{dice}}$ and binary cross-entropy loss $\mathcal{L}_{\text{bce}}$ between the predicted grounding masks and the ground truth masks of each phrase, following~\citet{cheng2022masked}. 
In cases where only box annotations are present, we apply the projection loss $\mathcal{L}_{\text{proj}}$ as introduced by \citet{tian2021boxinst}, which selects the mask whose projection on the axis matches the best with the annotated box. 
Essentially, this can be seen as a 1D dice loss calculated between the projected masks and the edges of the ground truth boxes along both the $x$ and $y$ axes. 
Given that the primary objective of grounding is to accurately select the correct mask, we assign different weights to these loss components. 
The mask dice loss and box projection loss are both weighted at 1, while the mask bce loss is given a lower weight of 0.1. 
The final loss calculation is a summation of the language modeling loss $\mathcal{L}_{\text{lm}}$ and these mask-related losses.

\paragraph{LLM Configuration}
We adopt the Vicuna-7B model~\cite{chiang2023vicuna} as our base LLM, and use the OpenAI CLIP@336~\cite{radford2021clip} model and DINOv2-L/14-reg\cite{darcet2023vision} pretrained checkpoints.
We use the original conversation template from Vicuna, where all the interactions are formatted as \texttt{<system\_message> <s> USER: <utterance> ASSISTANT: <utterance> </s>}.

\paragraph{Parameter-Efficient Training.}
We freeze all the parameters of the Mask2Former+, the CLIP, and the DINOv2 model during training. We use Low-Rank Adaptation (LoRA)~\cite{hu2021lora} with $r=16$ and $\alpha=16$ to tune the LLM, including all the linear layers, input embeddings, and the LM head.
We train all the new components introduced for connecting these models, including the MLP projection layer of CLIP and DINOv2, and the mask retrieval head. 
As a result, less than 2\% of the total parameters are trainable in the whole model.
We use the AdamW optimizer~\cite{loshchilov2017decoupled} with an initial learning rate of 2e-4 and a cosine annealing rate. We train our model on the balanced sampled M3G2 dataset for 2 epochs, which takes around 2 days using 8 40G A100 GPUs. 
\section{Additional Qualitative Results}

\begin{figure*}[h]
    \centering
    \includegraphics[width=1.0\linewidth]{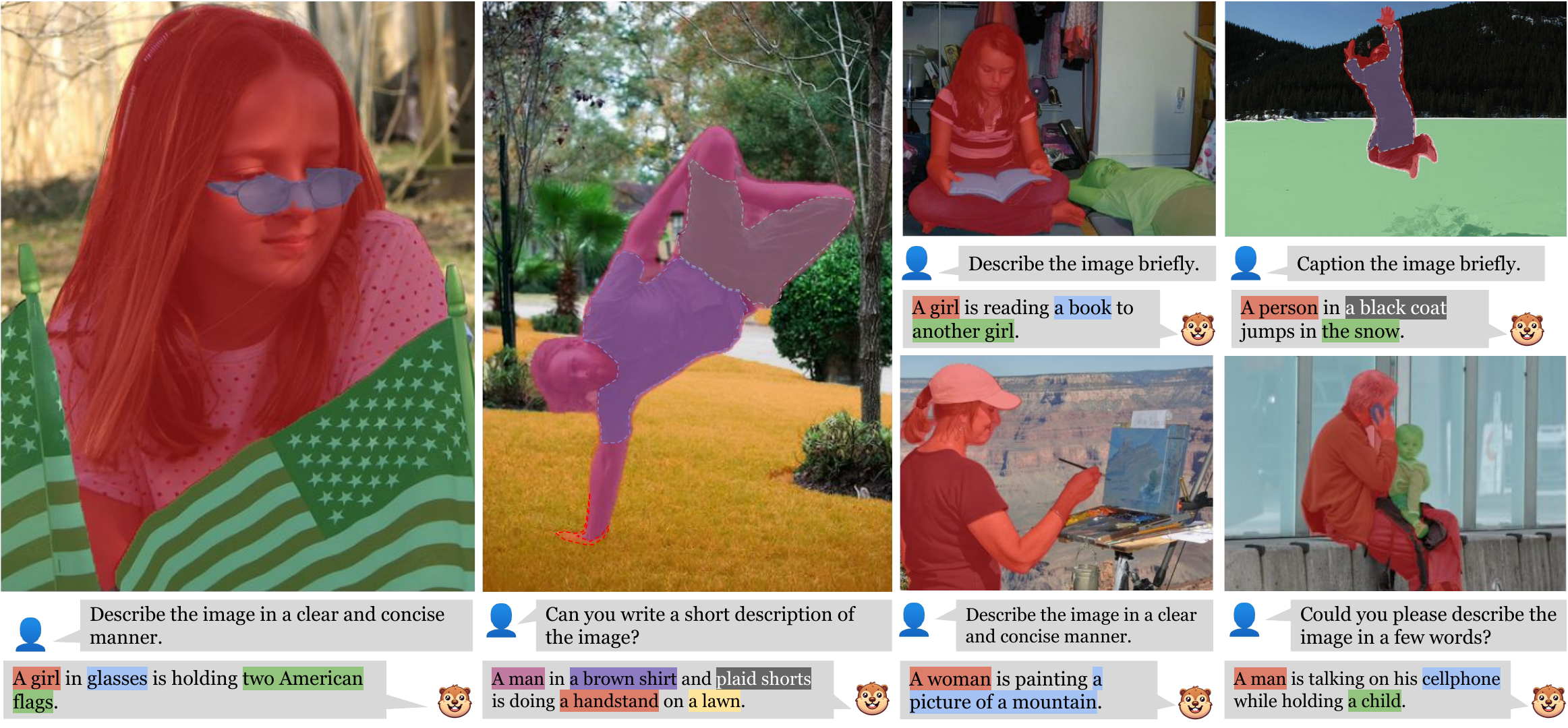}
    \vspace{-10pt}
    \caption{Additional qualitative examples for short grounded captioning.}
    \label{fig:qual-brief-cap}
\end{figure*}

\begin{figure*}[h]
    \centering
    \includegraphics[width=1.0\linewidth]{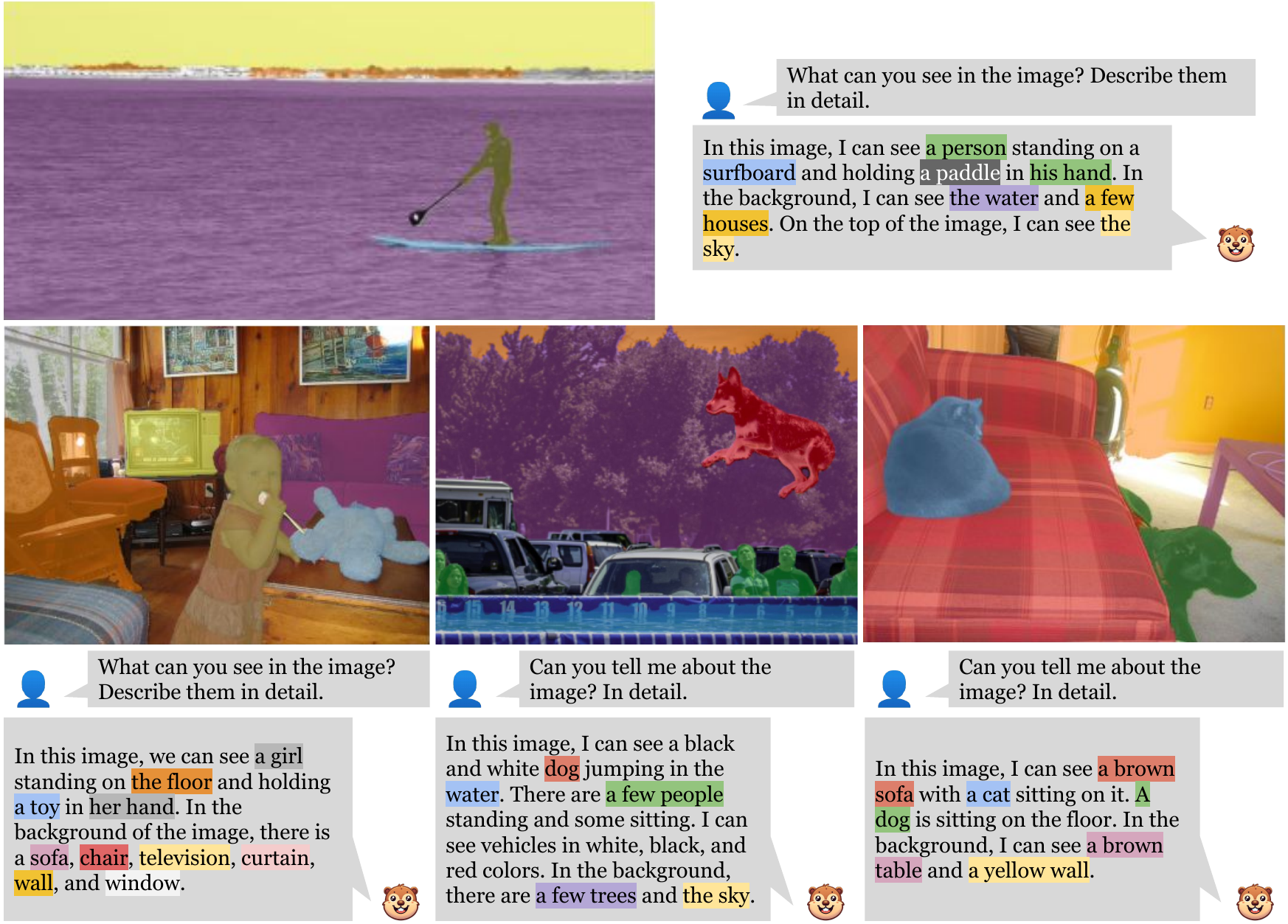}
    \vspace{-10pt}
    \caption{Additional qualitative examples for detailed grounded captioning.}
    \label{fig:qual-dense-cap}
\end{figure*}

\begin{figure*}[h]
    \centering
    \includegraphics[width=1.0\linewidth]{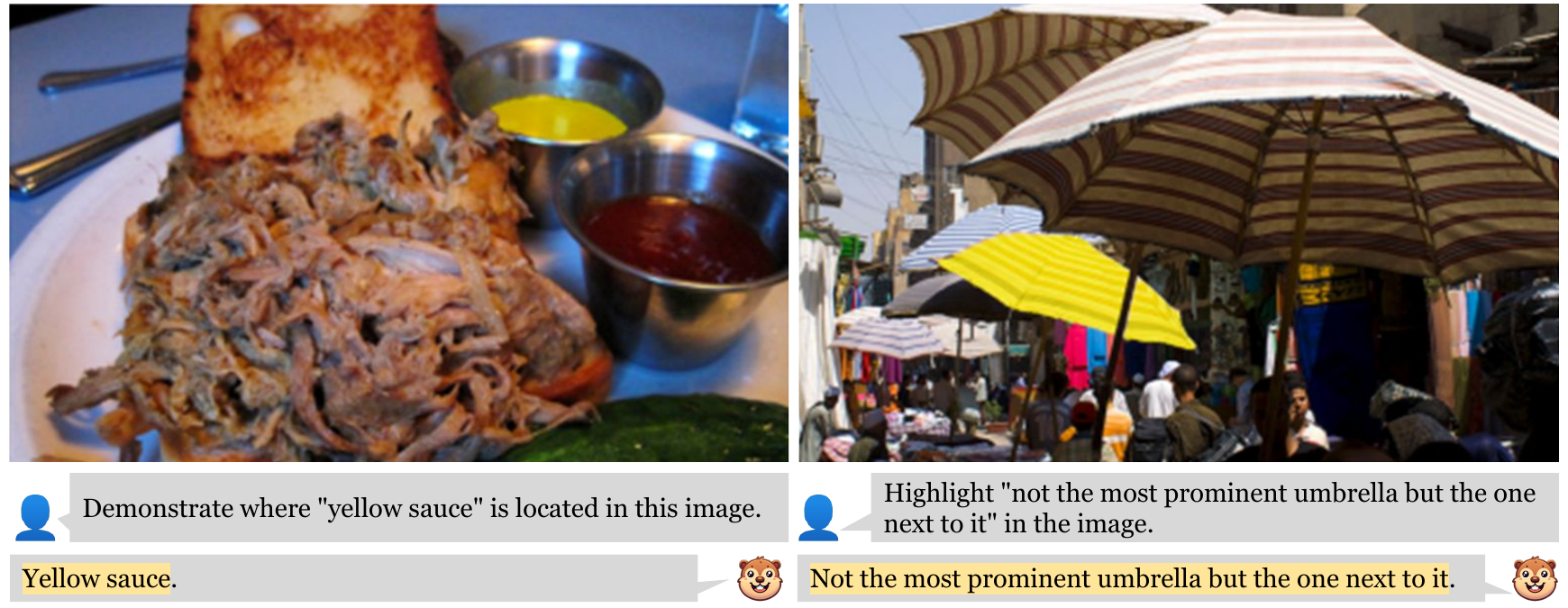}
    \vspace{-10pt}
    \caption{Additional qualitative examples for referring expression segmentation.}
    \label{fig:qual-res}
\end{figure*}

\begin{figure*}[h]
    \centering
    \includegraphics[width=1.0\linewidth]{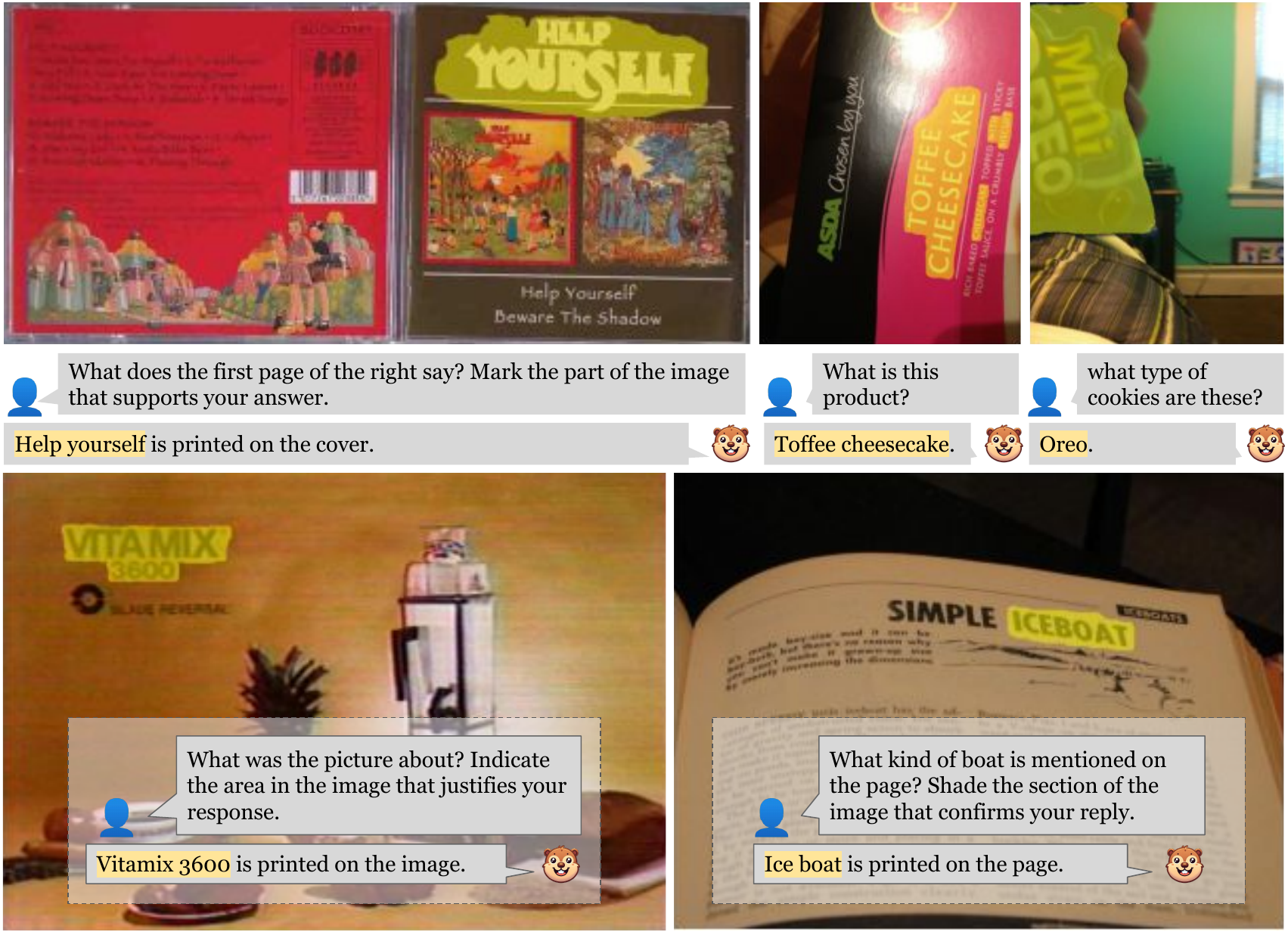}
    \vspace{-10pt}
    \caption{Additional qualitative examples for grounded VQA of visual text.}
    \label{fig:qual-textvqa}
\end{figure*}

\begin{figure*}[h]
    \centering
    \includegraphics[width=1.0\linewidth]{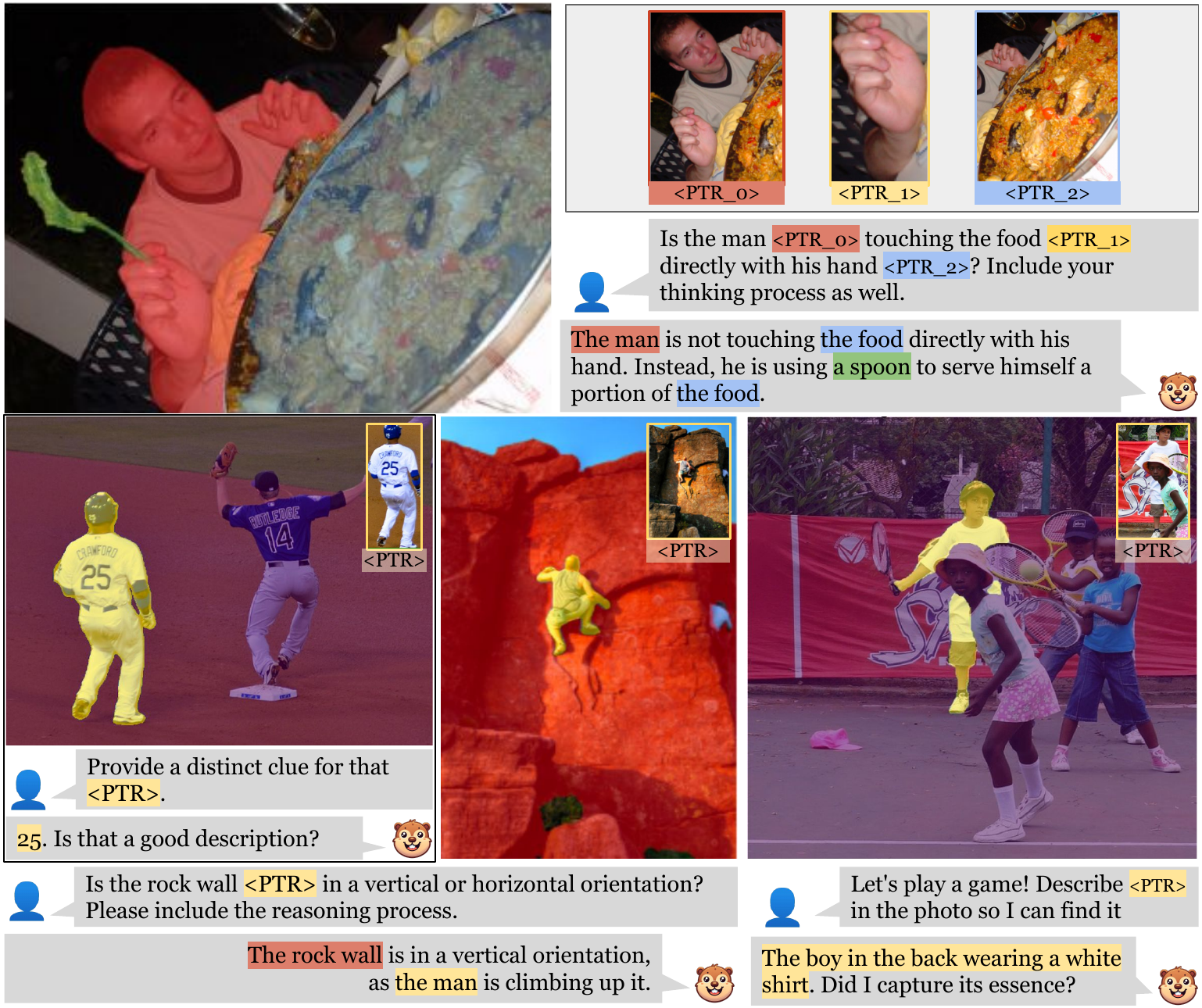}
    \vspace{-10pt}
    \caption{Additional qualitative examples for referential dialogue with pointer input.}
    \label{fig:qual-rd}
\end{figure*}

\begin{figure*}[h]
    \centering
    \includegraphics[width=1.0\linewidth]{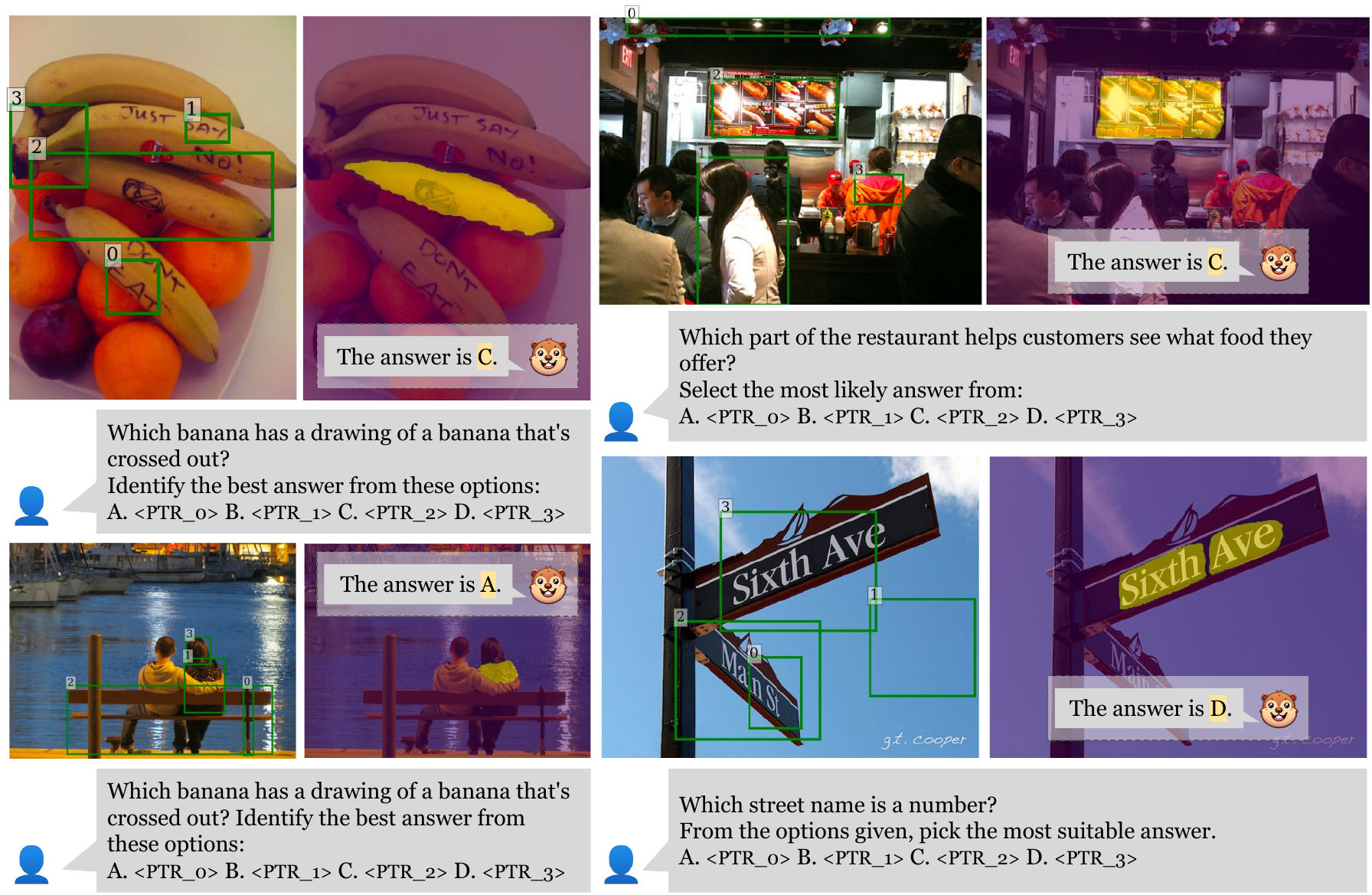}
    \vspace{-10pt}
    \caption{Additional qualitative examples for referential dialogue with pointers as multiple choices input.}
    \label{fig:qual-rd-mc}
\end{figure*}

We provide additional qualitative examples generated by \model. See Figure \ref{fig:qual-brief-cap} \& \ref{fig:qual-dense-cap} for GIC, Figure \ref{fig:qual-res} for RES, Figure \ref{fig:qual-textvqa} for GVQA, and Figure~\ref{fig:qual-rd} \& \ref{fig:qual-rd-mc} for RD, respectively. 

\begin{figure*}
\centering
\begin{subfigure}[t]{.3\textwidth}
\centering
\includegraphics[width=1.0\linewidth]{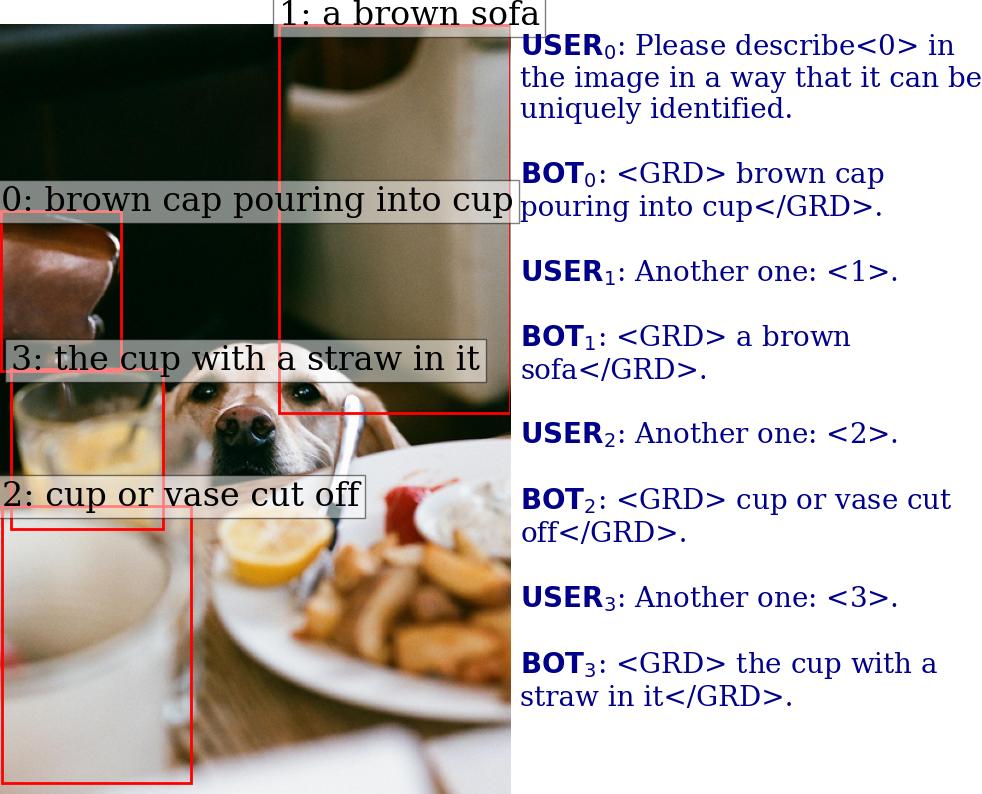}
\vspace{-10pt}
\caption{Refer expression generation (RefCOCO+).}
\label{fig:data_sample_refcoco+_2}
\end{subfigure}
~
\begin{subfigure}[t]{.33\textwidth}
\centering
\includegraphics[width=1.0\linewidth]{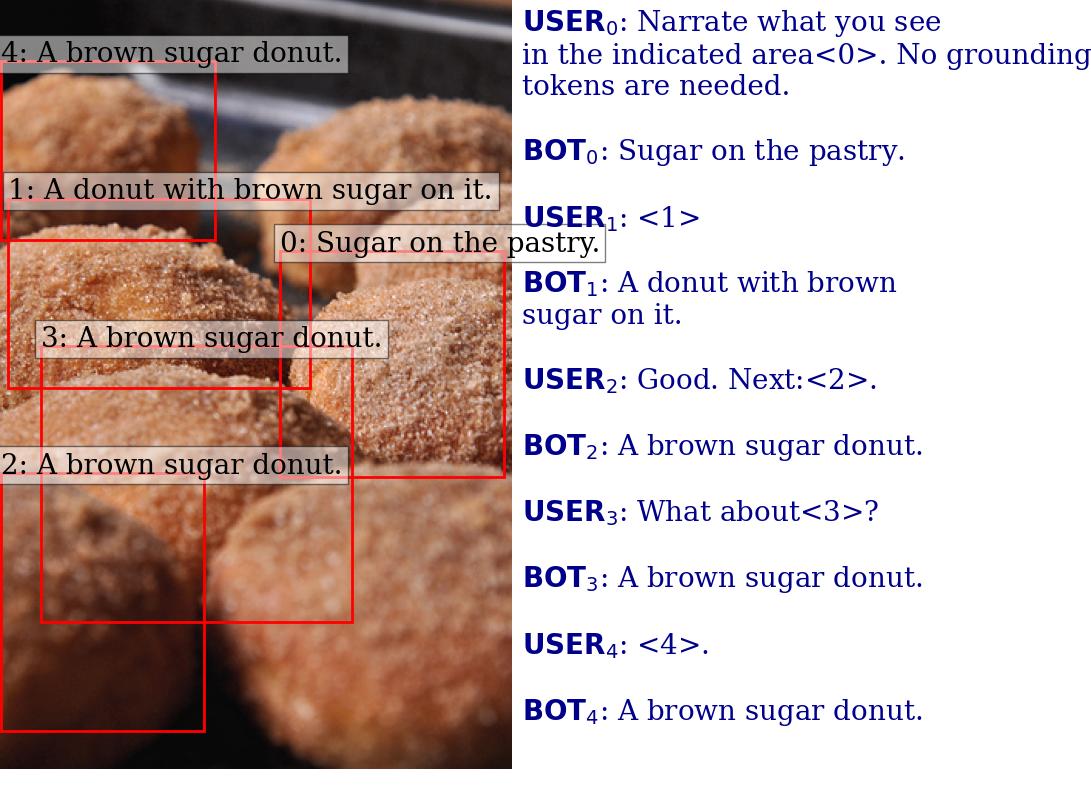}
\vspace{-10pt}
\caption{Region captioning (VG).}
\label{fig:data_sample_vg_1}
\end{subfigure}
~
\begin{subfigure}[t]{.34\textwidth}
\centering
\includegraphics[width=1.0\linewidth]{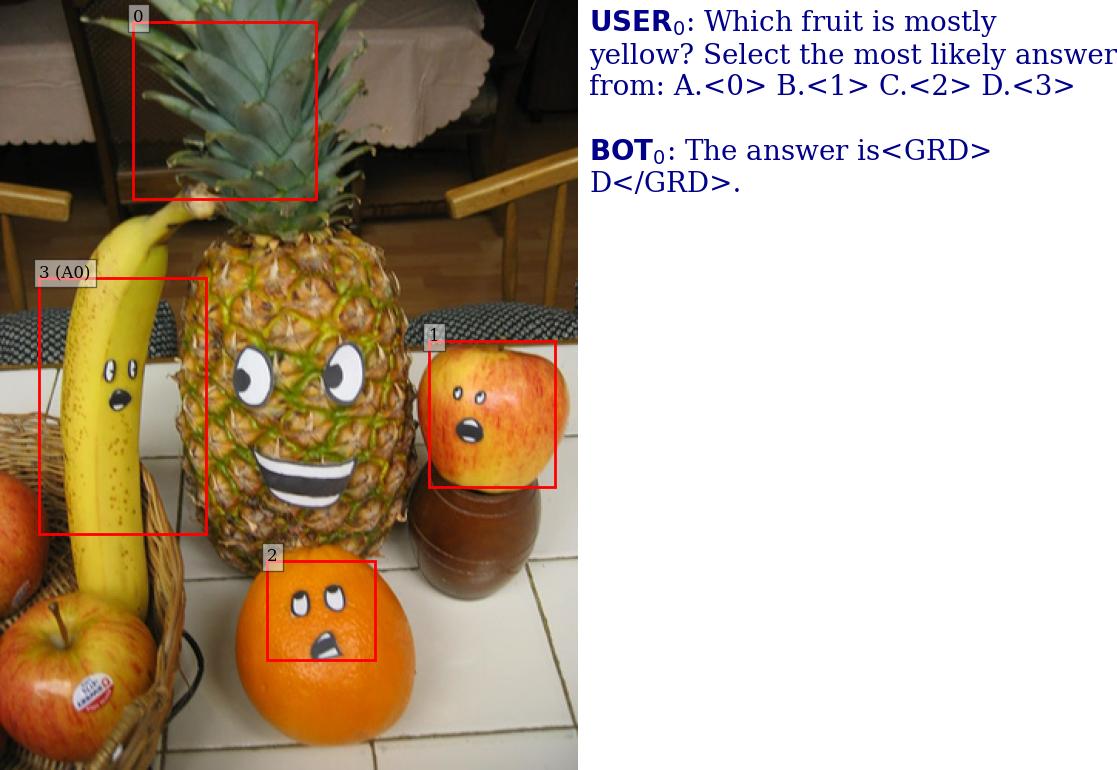}
\vspace{-10pt}
\caption{Example of the referring QA task from V7W.}
\label{fig:data_sample_v7w_1}
\end{subfigure}
~
\begin{subfigure}[t]{.48\textwidth}
\centering
\includegraphics[width=.9\linewidth]{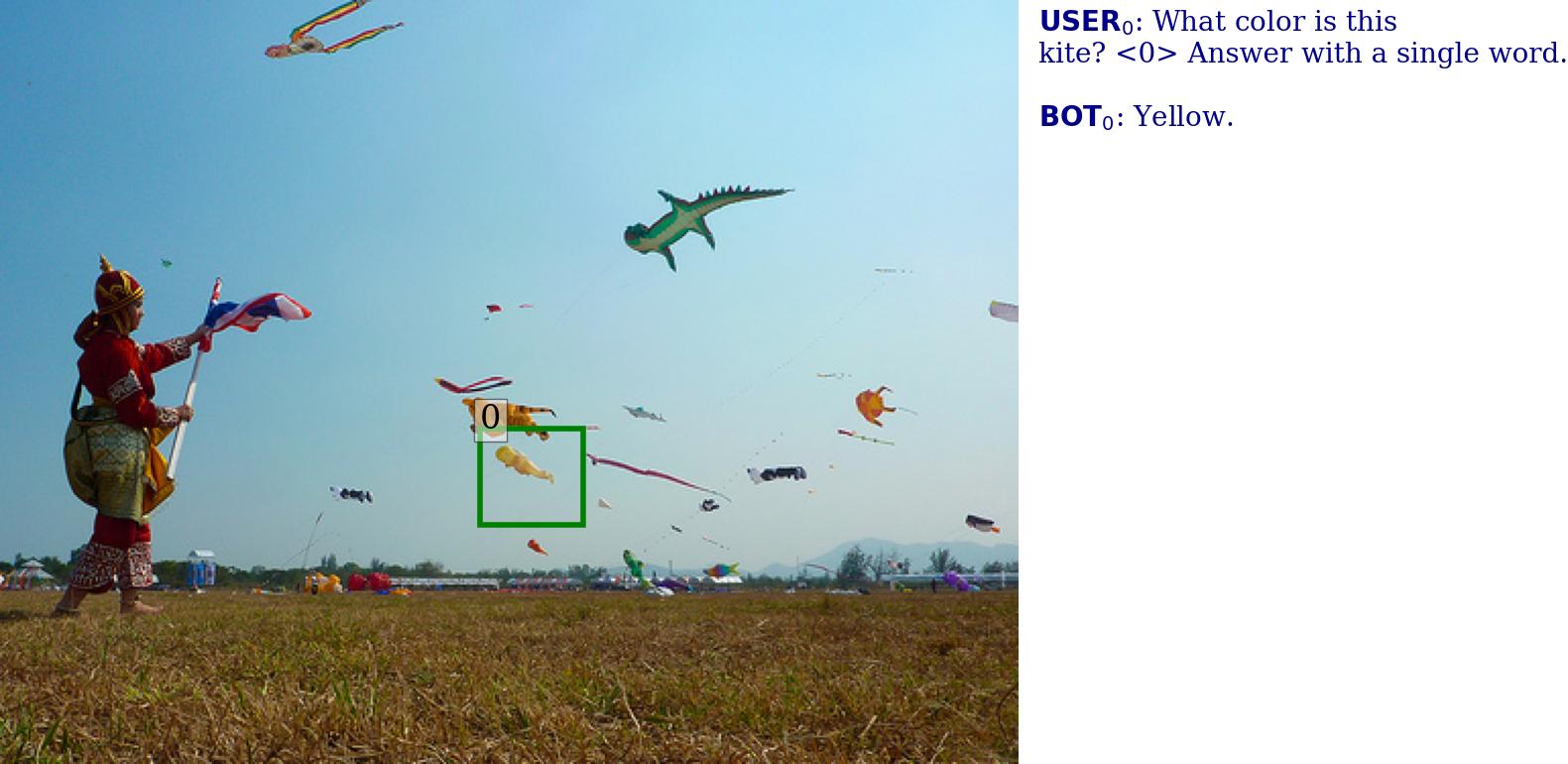}
\caption{Example of the referring QA task originated from PointQA-Local.}
\label{fig:data_sample_pointqa_1}
\end{subfigure}
~
\begin{subfigure}[t]{.48\textwidth}
\centering
\includegraphics[width=.9\linewidth]{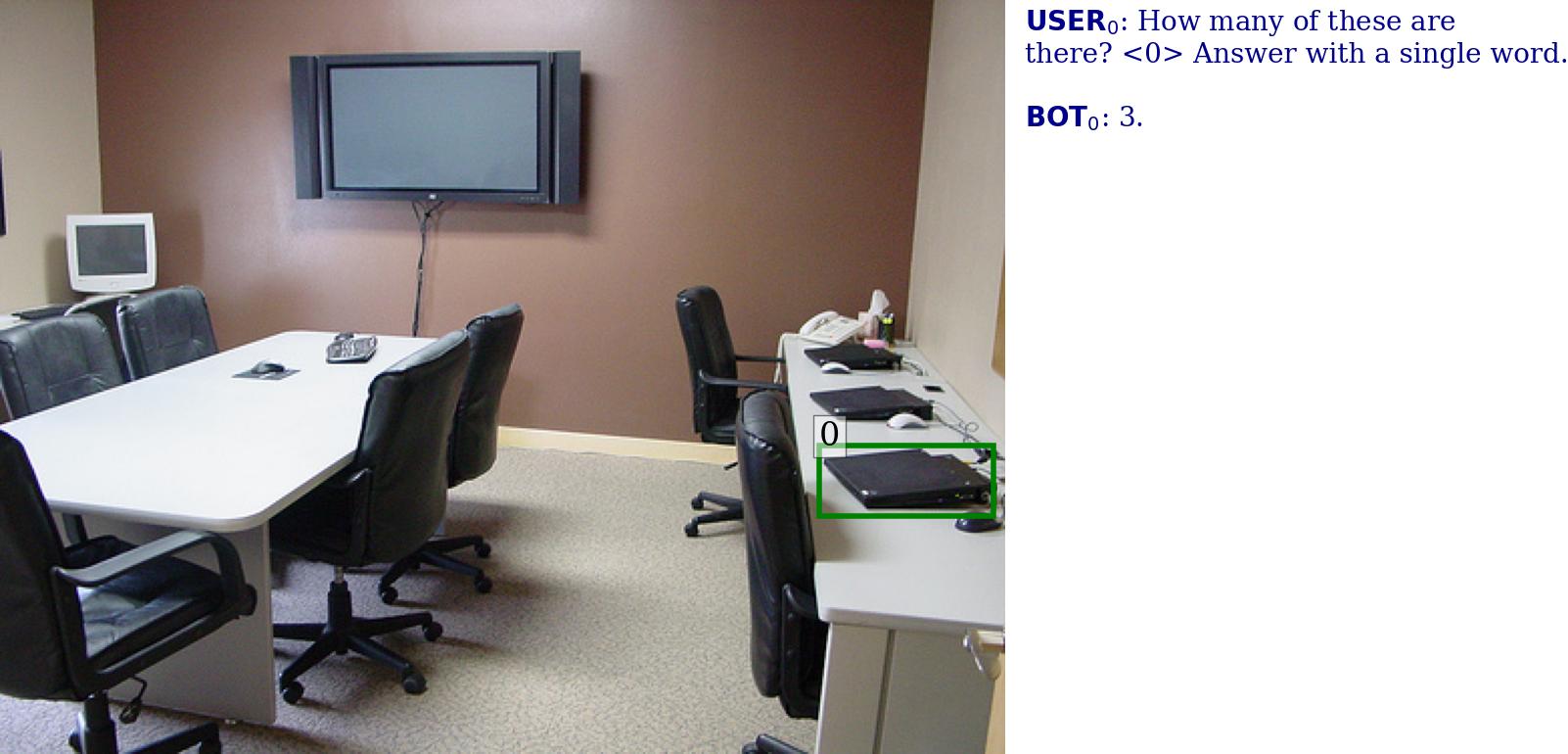}
\caption{Example of the referring QA task originated from PointQA-Twice.}
\label{fig:data_sample_pointqa_2}
\end{subfigure}
~
\begin{subfigure}[t]{1.0\textwidth}
\centering
\includegraphics[width=1.0\linewidth]{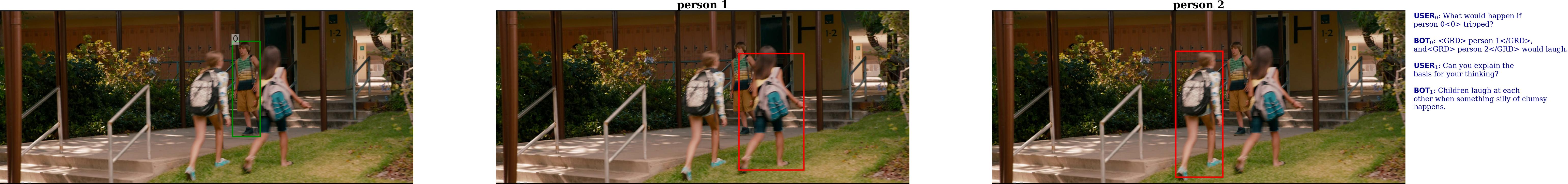}
\vspace{-10pt}
\caption{Example of the referring QA task originated from VCR.}
\label{fig:data_sample_vcr_1}
\end{subfigure}
~
\begin{subfigure}[t]{.97\textwidth}
\centering
\includegraphics[width=1.0\linewidth]{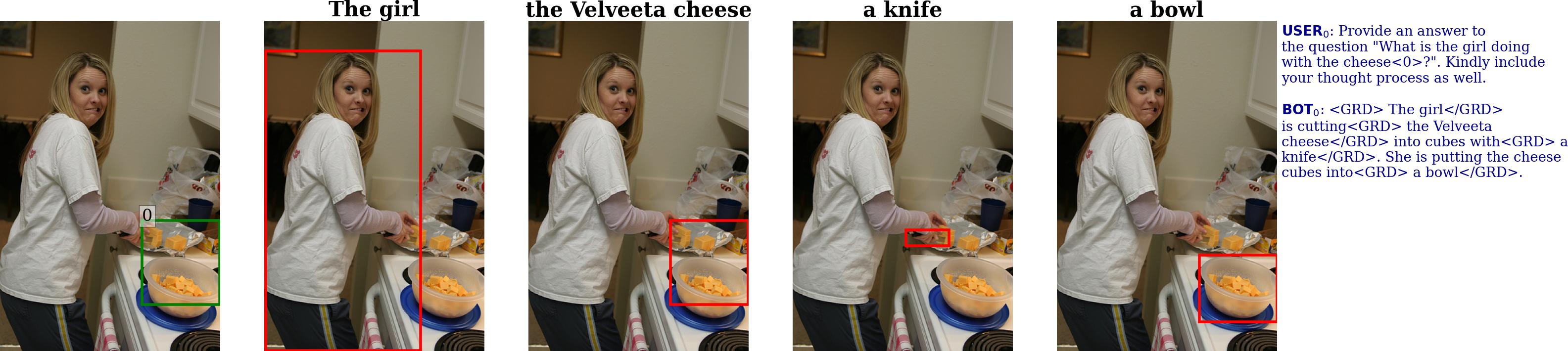}
\vspace{-10pt}
\caption{Example of the referential dialog task originated from ShikraRD.}
\label{fig:data_sample_shikra_2}
\end{subfigure}
~
\begin{subfigure}[t]{.97\textwidth}
\centering
\includegraphics[width=1.0\linewidth]{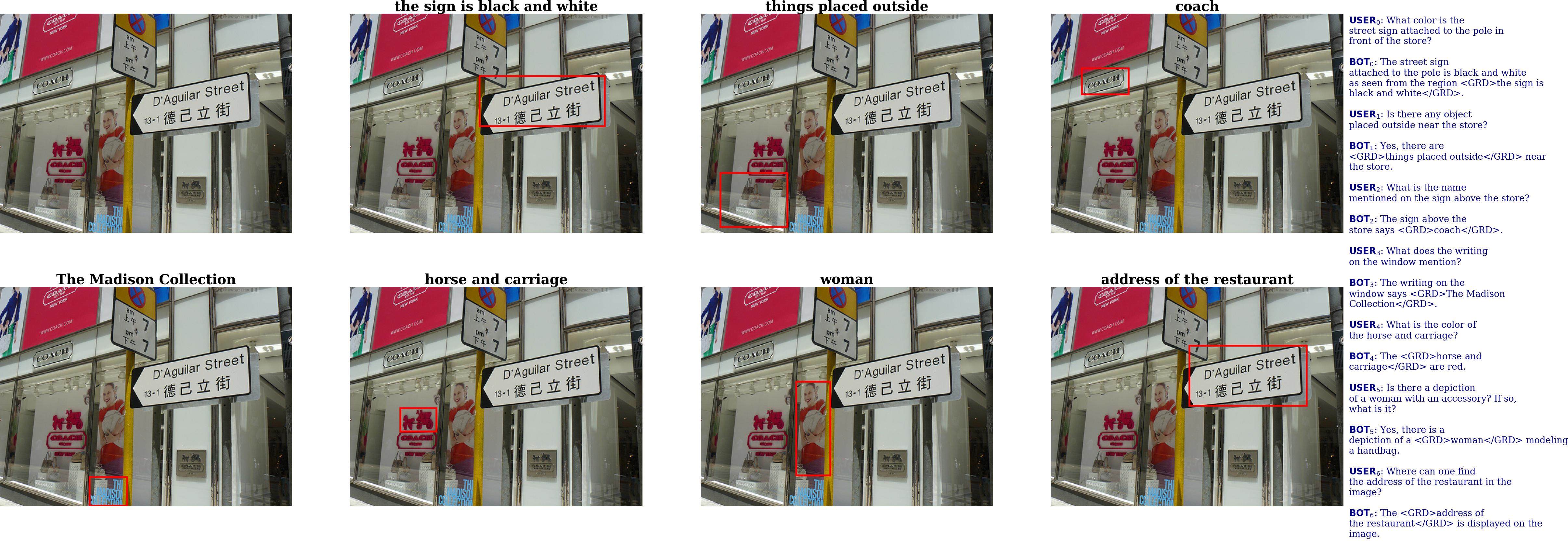}
\vspace{-30pt}
\caption{Example of the referential dialog task originated from SVIT.}
\label{fig:data_sample_svit_1}
\end{subfigure}
\vspace{-5pt}
\caption{Data samples of the RD task (part 1).}
\label{fig:data_sample_rd1}
\end{figure*}

\end{document}